\documentclass[a4paper,fleqn]{cas-sc}

\usepackage[numbers]{natbib}

\usepackage {color}
\usepackage{soul}

\usepackage{lineno}
\def\tsc#1{\csdef{#1}{\textsc{\lowercase{#1}}\xspace}}
\tsc{WGM}
\tsc{QE}

\begin{document}
\let\printorcid\relax
\let\WriteBookmarks\relax
\def\floatpagepagefraction{1}
\def\textpagefraction{.001}
\shorttitle{{A Survey on Neural-symbolic Learning Systems} }   

\shortauthors{Dongran Yu et al.}  

\title [mode = title]{{A Survey on Neural-symbolic Learning Systems} }  

\tnotemark[<tnote number>] 


%

\author[1,3]{Dongran Yu}[style=chinese]



\ead{yudran@foxmail.com}


\credit{Investigation, Data curation, Methodology and Writing}

\address[1]{School of Computer Science and Technology and the Key Laboratory of Symbolic Computation and Knowledge Engineer (Jilin University), Ministry of Education, Changchun, Jilin 130012, China }

\author[1,2]{Bo Yang}[ style=chinese]


\ead{ybo@jlu.edu.cn}


\credit{Conceptualization,Supervision, Review and Editing}

\address[2]{School of Computer Science and Technology, Jilin University, Changchun, Jilin, 130012, China}
\cormark[1]

\author[1,2]{Dayou Liu}[ style=chinese]
\credit{Supervision}
\address[3]{School of Artificial Intelligence, Jilin University, Changchun, Jilin, 130012, China}
\author[4]{Hui Wang}[ style=chinese]
\address[4]{School of Electronics, Electrical Engineering and Computer Science, Queen's University Belfast}
\credit{Supervision}

\author[5]{Shirui Pan}[ style=chinese]
\ead{s.pan@griffith.edu.au}
\address[5]{School of Information and Communication Technology, Griffith University}
\credit{Supervision, Review and Editing}
\cortext[1]{Corresponding author. School of Computer Science and Technology, Jilin University, Changchun, Jilin, 130012, China (Note:Accepted by Neural Networks Journal)}



\begin{abstract}
In recent years, neural systems have demonstrated highly effective learning ability and superior perception intelligence. However, they have been found to lack effective reasoning and cognitive ability. On the other hand, symbolic systems exhibit exceptional cognitive intelligence but suffer from poor learning capabilities when compared to neural systems. Recognizing the advantages and disadvantages of both methodologies, an ideal solution emerges: combining neural systems and symbolic systems to create neural-symbolic learning systems that possess powerful perception and cognition. The purpose of this paper is to survey the advancements in neural-symbolic learning systems from four distinct perspectives: challenges, methods, applications, and future directions. By doing so, this research aims to propel this emerging field forward, offering researchers a comprehensive and holistic overview. This overview will not only highlight the current state-of-the-art but also identify promising avenues for future research.
 
\end{abstract}


\begin{highlights}
\item Neural-symbolic learning systems combine the neural systems and the symbolic systems into a unified framework.
\item Neural-symbolic learning systems can equip AI with the ability to perform perception and cognition.
\item A good combination between the neural systems and the symbolic systems allows the model to achieve the desired performance.
\end{highlights}

\begin{keywords}
 Neural-symbolic Learning Systems\sep Neural Networks\sep Symbolic Reasoning\sep Symbols\sep Logic\sep Knowledge Graphs
\end{keywords}

\maketitle
\section{Introduction}
\begin{figure}[!t]
\centering
\includegraphics[width=0.8\linewidth]{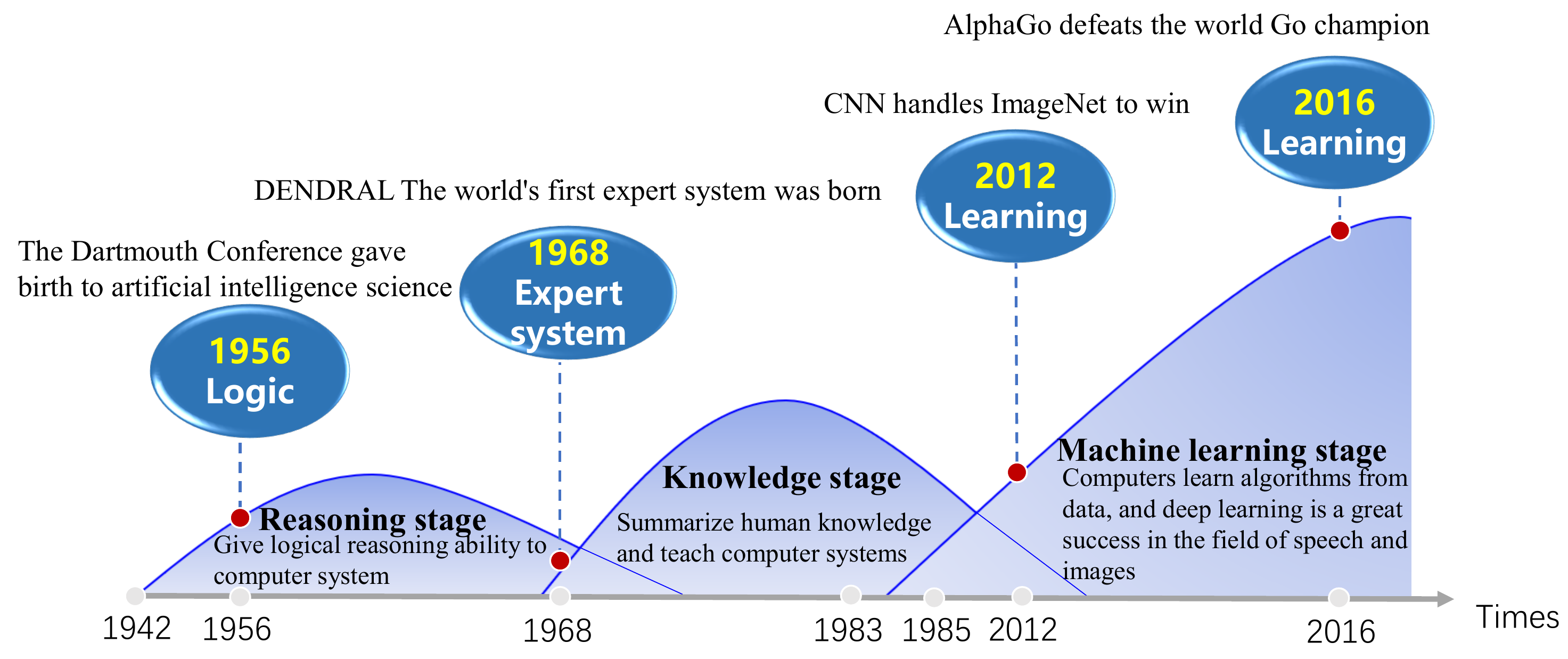}
\caption{The brief history of artificial intelligence.}
\label{history}
\end{figure}

Perception, represented by connectionism or neural systems, and cognition, represented by symbolism or symbolic systems, are two fundamental paradigms in the field of artificial intelligence (AI), each having prevailed for several decades. Figure \ref{history} showcases the rise and fall of these two doctrines, which researchers commonly categorize into three significant periods. (1) The 1960s-1970s (1956-1968). The inception of reasoning in AI can be traced back to the 1956 Dartmouth Conference, where Newell and Simon introduced the "logical theorist" program, successfully proving 38 mathematical theorems. This marked the beginning of the reasoning era in AI. However, researchers soon realized that relying solely on heuristic search algorithms had limitations, and many complex problems necessitated specialized domain knowledge to achieve higher levels of intelligence. Consequently, incorporating knowledge into AI models became a prevalent notion, as it enhanced the ability to find solutions within large solution spaces. (2) The 1970s-1990s (1968-1985). During this period, significant developments occurred, such as the creation of the first expert system, "DenDral" by Faigan's nanny and Lederberg in 1968. This milestone represented the organic integration of AI and domain knowledge, marking the advent of the knowledge-focused era in AI. However, knowledge acquisition posed a significant challenge, leading to a shift towards automatic acquisition of valuable knowledge from massive datasets, which gradually became the mainstream trend in AI. (3) From the 1990s to the present: In 1983, neural networks started gaining prominence, and after 2000, AI entered the era of machine learning. Notable breakthroughs in machine learning, particularly through neural networks, include the 2012 triumph of ImageNet with Deep Convolutional Neural Networks (CNN), and the 2016 victory of AlphaGo against the Go world champion. These significant milestones exemplify the power of machine learning approaches, predominantly driven by neural networks.

To date, neural networks have demonstrated remarkable accomplishments in perception-related tasks, such as image recognition \cite{9165588}. However, there exist various scenarios, including question answering \cite{gupta2022vquad}, medical diagnosis \cite{salahuddin2022transparency}, and autonomous driving \cite{wen2022deep}, where relying solely on perception can present limitations or yield unsatisfactory outcomes. For instance, when confronted with unseen situations during training, machines may struggle to make accurate decisions in medical diagnosis. Another crucial consideration is the compatibility of purely perception-based models with the principles of explainable AI  \cite{ratti2022explainable}. Neural networks, being black-box systems, are unable to provide explicit calculation processes. In contrast, symbolic systems offer enhanced appeal in terms of reasoning and interpretability. For example, through deductive reasoning and automatic theorem proving, symbolic systems can generate additional information and elucidate the reasoning process employed by the model.

Consequently, an increasing number of researchers have directed their attention towards the fusion of neural systems and symbolic systems, aiming to achieve the third wave of AI: neural-symbolic learning systems \cite{mao2019neuro,manhaeve2018deepproblog,dong2019neural,marra2019integrating,zhou2019abductive,evans2018learning}. In a special NeurIPS 2019 lecture, Turing Award laureate Yoshua Bengio drew inspiration from Dr. Daniel Kahneman's renowned book "Thinking Fast and Slow" \cite{kahneman2011thinking} to emphasize the need for a system-1-to-system-2 transformation in deep learning. Here, system 1 represents the intuitive, rapid, unconscious, nonlinguistic, and habitual aspects, while system 2 embodies the deliberative, logical, sequential, conscious, linguistic, algorithmic, planning-related, and reasoning-related facets. Indeed, since the 1990s, numerous researchers in the fields of artificial intelligence and cognitive science have explicitly proposed the concept of dual processes that correspond to these contrasting systems \cite{honavar1995symbolic}. These highlight the necessity of combining neural systems and symbolic systems. By unifying these two system types within a comprehensive framework, \textbf{neural-symbolic learning systems} can be created, endowing AI with the capability to perform both perception and reasoning tasks. It is worth noting that the idea of integrating neural systems and symbolic systems, referred to as hybrid connectionist-symbolic models, was initially introduced in the 1990s \cite{sun1994computational}.

\begin{table*}[h]
\renewcommand{\arraystretch}{1.3}
\centering
\caption{Summarize properties for the symbolic systems and neural systems separately.}
\label{1}
\newcommand{\tabincell}[2]{\begin{tabular}{@{}#1@{}}#2\end{tabular}}
\resizebox{\textwidth}{!}{
\begin{tabular}{c|c|c|c|c|c}

\hline
\textbf{Systems} & \textbf{Processing Methods}&\textbf{Knowledge representation}&\textbf{Primary algorithms}&\textbf{Advantages} & \textbf{Disadvantages}  \\ 
\hline
\multirow{3}*{\tabincell{c}{Symbolic systems }}&&&&Strong generalization ability& Weak at handling unstructured data 
\\
&Deductive reasoning&Logical representation&Logical deduction&Good interpretability& Weak robustness  \\
 
&&&&Knowledge-driven& Slow reasoning \\ \hline
\multirow{3}*{\tabincell{c}{Neural systems\\ (Sub-symbolic systems)}}&&&& Strong at handling unstructured data& Weak generalizability (adaptability) \\

&Inductive learning&Distributed representation&BP algorithms& Strong robustness & Lack of interpretability  \\

 &&&&Fast learning & Data-driven    \\ \hline
\end{tabular}}
\end{table*}

Neural-symbolic learning systems leverage the combine strengths of both neural systems and symbolic systems \cite{manhaeve2019deepproblog,zuidberg2020symbolic,perotti2012learning,garcez2015neural,towell1994knowledge,besold2017neural,garcez2012neural,kaur2017relational,sourek2018lifted,garcez1999connectionist,dragone2021neuro,karpas2022mrkl,landajuela2021discovering,levine2022standing}. To provide a comprehensive understanding, the survey initially outlines key characteristics of symbolic systems and neural systems  (refer to Table \ref{1}), including processing methods, knowledge representation, etc. Analysis of Table \ref{1} reveals that symbolic systems and neural systems exhibit complementary features across various aspects. For instance, symbolic systems may possess limited robustness, whereas neural systems demonstrate robustness. Consequently, neural-symbolic learning systems emerge as a means to compensate for the shortcomings inherent in individual systems.

\begin{figure}[!t]
\centering
\includegraphics[width=0.6\linewidth]{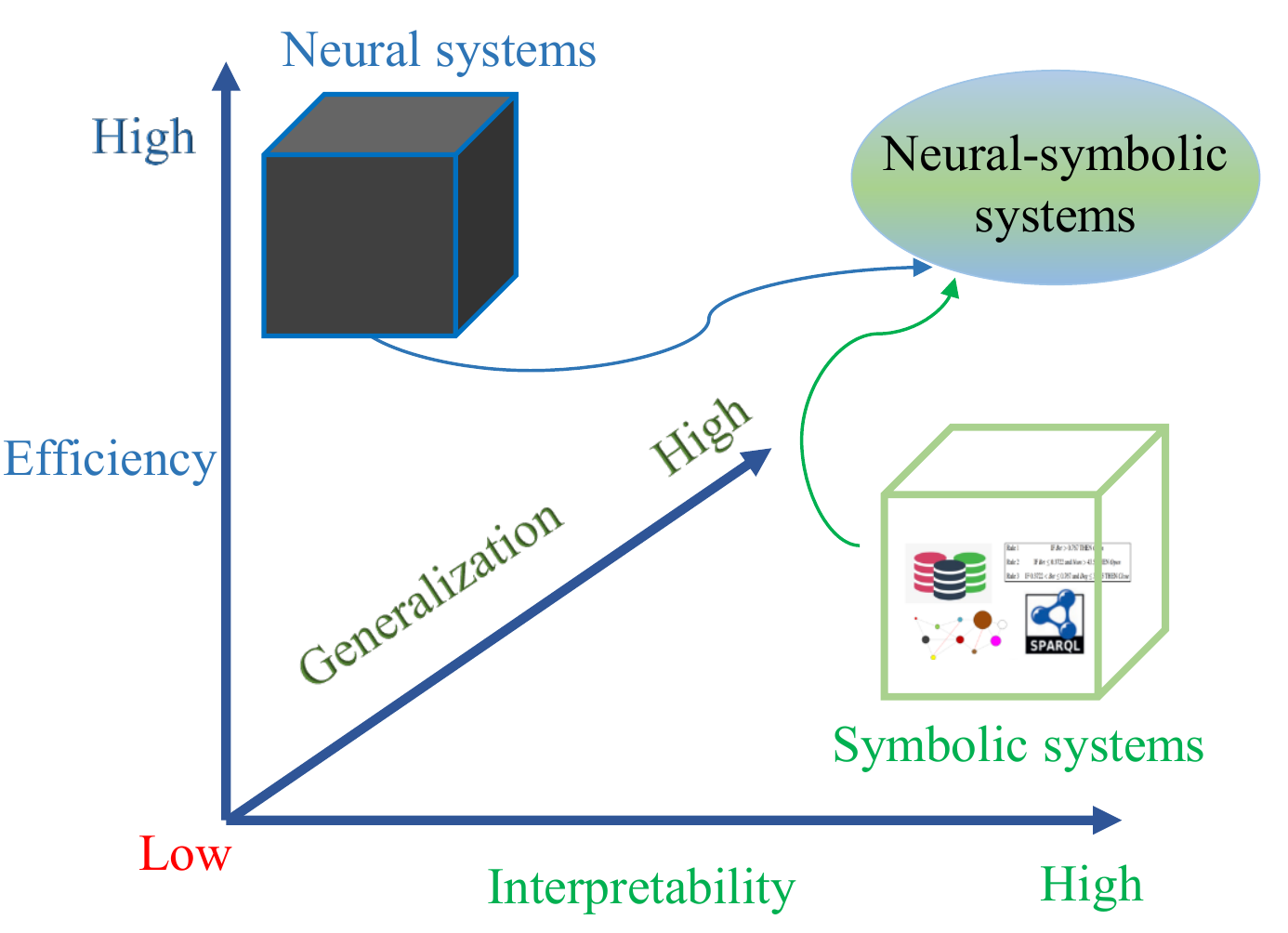}
\caption{The advantages of neural-symbolic learning systems with respect to model efficiency, generalization, and interpretability. The neural systems are black-box systems, while symbolic systems are white-box systems.}
\label{fig:2}
\end{figure}

Moreover, we conduct an analysis of neural-symbolic learning systems from three key perspectives: efficiency, generalization, and interpretability. As depicted in Figure \ref{fig:2}, neural-symbolic learning systems excel in these areas. Firstly, in terms of efficiency, neural-symbolic learning systems can reason quickly compared to pure symbolic systems, thereby reducing computational complexity \cite{zhang2020efficient}. This accelerated computation can be attributed to the integration of neural networks, as outlined in Section \ref{sec3} ({\em Learning for reasoning}). Traditional symbolic approaches typically employ search algorithms to navigate solution spaces, leading to increased computational complexity as the search space grows in size. Secondly, with regard to generalization, neural-symbolic learning systems outperform standalone neural systems in terms of their capacity for generalization. The incorporation of symbolic knowledge as valuable training data enhances the model's generalization abilities \cite{kampffmeyer2019rethinking} (see Section \ref{sec3} {\em Reasoning for learning}). Thirdly, in terms of interpretability, neural-symbolic learning systems represent gray-box systems, in contrast to standalone neural systems. By leveraging symbolic knowledge, these systems can provide explicit computation processes, such as traced reasoning processes or chains of evidence for results \cite{yang2019learn}. Consequently, neural-symbolic learning systems have emerged as vital components of explainable AI, yielding superior performance across diverse domains, including computer vision  \cite{xie2019embedding,donadello2017logic,wang2018zero,kampffmeyer2019rethinking,zhu2014reasoning,chen2020knowledge,li2019large}, and natural language processing \cite{liang2016neural,yi2018neural,tian2022weakly}, etc.

\textbf{Challenge:} Symbolic systems and neural systems diverge in terms of their data representations and problem-solving approaches. Symbolic systems rely on discrete symbolic representations and traditional search algorithms to discover solutions, while neural systems employ continuous feature vector representations and neural cells to learn mapping functions. Consequently, a significant challenge lies in designing a unified framework that seamlessly integrates both symbolic and neural components. The aim is to strike a balance and select an appropriate combination of symbolic and neural systems that aligns with the requirements of the specific problem \cite{honavar1995symbolic}.

To provide new readers with a comprehensive understanding of neural-symbolic learning systems, this paper surveys representative research and applications of these systems. To summarize and systematically review related works, there are several surveys conducted during the past few years \cite{andrews1995survey,garcez2002neural, besold2017neural, calegari2020integration,8889997, lamb2020graph, marra2021statistical,von2021informed,townsend2019extracting,garcez2020neurosymbolic,sun1994computational, sun2013connectionist}, \cite{kautz2022third,marcus2020next}. 
For example, \cite{andrews1995survey,townsend2019extracting} center around knowledge extraction techniques, which aligns with the first category discussed in Section \ref{sec3}. On the other hand, \cite{lamb2020graph,calegari2020integration,von2021informed,marra2021statistical} provide detailed reviews from specific perspectives, such as graph neural networks (GNNs), prior knowledge integration, explainable artificial intelligence (XAI), and statistical relational learning. While surveys \cite{besold2017neural,garcez2020neurosymbolic} also cover neural-symbolic learning systems comprehensively, their focus remains primarily theoretical, lacking a thorough introduction to specific techniques and related works. Therefore, an urgent need arises to provide a comprehensive survey that encompasses popular methods and specific techniques (e.g., model frameworks, execution processes) to expedite advancements in the neural-symbolic field. Distinguishing itself from the aforementioned surveys, this paper emphasizes classifications, techniques, and applications within the domain of neural-symbolic learning systems.

\textbf{Motivation:} For our part, we do not seek to replace the above literature but complement them by offering a comprehensive overview of the broader domain of neural-symbolic learning systems. This encompasses various technologies, cutting-edge developments, and diverse application areas within the realm of neural-symbolic learning systems. Additionally, this article caters to individuals engaged in the pursuit of integrating symbolic systems and neural systems. With an emphasis on integration, we present a novel classification framework for neural-symbolic learning systems in Section \ref{sec3}.

Our contributions can be summarized as follows:

1) We propose a novel taxonomy of neural-symbolic learning systems. Neural-symbolic learning systems are categorized into three groups: {\em learning for reasoning}, {\em reasoning for learning}, and {\em learning\mbox{-}reasoning}.

2) We provide a comprehensive overview of neural-symbolic techniques, along with types and representations of symbols such as logic knowledge and knowledge graphs. For each taxonomy, we provide detailed descriptions of the representative methods, summarize the corresponding characteristics, and give a new understanding of neural-symbolic learning systems.

3) We discuss the applications of neural-symbolic learning systems and propose four potential future research directions, thus paving the way for further advancements and exploration in this field.

The remainder of this survey is organized as follows. In Section \ref{sec3}, we categorize the different methods of neural-symbolic learning systems. Section \ref{sec4} introduces the main technologies of neural-symbolic learning systems. We summarize the main applications of neural-symbolic learning systems in Section \ref{sec5}. Section \ref{sec6} discusses the future research directions, after which Section \ref{sec7} concludes this survey.
\hfill 
\hfill

\begin{table*}[pos=!h]
\Huge
\renewcommand{\arraystretch}{2}
\centering
\caption{Main Approaches of neural-symbolic learning systems. Taxonomies of the approaches and combination modes are presented in Section \ref{sec3}. Methodical details can be found in Section \ref{sec4}. The symbols and neural networks used herein are introduced in Appendix \ref{sec2}, and "Agnostic" means arbitrary neural networks. "Serialization", "parallelization" and "interaction" are combination modes, respectively. Applications are discussed in Section \ref{sec6}.}
\label{tab:2}
\newcommand{\tabincell}[2]{\begin{tabular}{@{}#1@{}}#2\end{tabular}}
\renewcommand{\multirowsetup}{\centering}
\resizebox{\textwidth}{!}{
\begin{tabular}{|c|c|c|c|c|c|c|}

\hline
\textbf{Representative works} & \textbf{Taxonomies} & \textbf{Methods} & \textbf{Symbols} & \textbf{Neural networks} & \textbf{Applications}  \\ 
\hline
pLogicNet\cite{qu2019probabilistic}& \multirow{6}*{\tabincell{c}{{\em Learning for reasoning }\\ (Serialization)}} & \multirow{5}*{\tabincell{c}{Accelerating}} & \multirow{8}*{\tabincell{c}{First\mbox{-}order logic}} &\multirow{2}*{\tabincell{c}{GNN}}& \multirow{2}*{Knowledge graph reasoning}
\\
\cline{0-0}
ExpressGNN\cite{zhang2020efficient}&&&&&
\\ \cline{6-6}
\cline{0-0}\cline{5-5}
RNM\cite{marra2020relational}&&& &CNN&Classification\\\cline{6-6}
\cline{0-0}\cline{5-5}
NMLN\cite{marra2019neural}&&&&GNN&\tabincell{c}{Knowledge graph reasoning}\\\cline{6-6}
\cline{0-0}\cline{5-5}
NLIL\cite{yang2019learn}&&&&CNN, GNN& \tabincell{c}{Classification  and \\knowledge graph reasoning}\\
\cline{0-0}\cline{3-3}\cline{6-6}\cline{5-5}
NS-CL\cite{mao2019neuro}&&Absracting&&RNN&Visual question answering\\
\cline{0-2}\cline{6-6}\cline{5-5}
HDNN\cite{hu2016harnessing}&\multirow{10}*{\tabincell{c}{{\em Reasoning for learning} \\ (Parallelization)}} &\multirow{8}*{Reguariziting}& & Agnostic& \tabincell{c}{ \tabincell{c}{Object recognition } }\\
\cline{0-0}\cline{6-6}\cline{5-5}
SBR\cite{diligenti2017semantic}& &  &&CNN&\multirow{2}*{Classification}\\ \cline{0-0}\cline{4-4}\cline{5-5}
SL\cite{xu2018semantic}&&&\multirow{2}*{\tabincell{c}{Propositional logic}}&Agnostic&\\
\cline{6-6}
\cline{0-0}\cline{5-5}
LENSR\cite{xie2019embedding}&&& &\multirow{3}*{\tabincell{c}{CNN,GNN}} & Visual relationship detection\\
\cline{0-0}\cline{4-4}\cline{6-6}
CA\mbox{-}ZSL\cite{luo2020context}&&&\multirow{5}*{\tabincell{c}{Knowledge graph }} & &\multirow{5}*{\tabincell{c}{Few-shot classification}} \\
\cline{0-0}
LSFSL\cite{li2019large}&&&&&\\
\cline{0-0}\cline{3-3}\cline{5-5}
SEKB\mbox{-}ZSL\cite{wang2018zero}&&\multirow{3}*{\tabincell{c}{Transfering}} &&GNN&\\
\cline{0-0}\cline{5-5}
DGP\cite{kampffmeyer2019rethinking}&&&&\multirow{2}*{\tabincell{c}{CNN,GNN}}& \\
\cline{0-0}
KGTN\cite{chen2020knowledge}&&&&&\\
\cline{0-0}\cline{4-4}\cline{6-6}\cline{5-5}
PROLONETS\cite{silva2021encoding}&&&Propositional logic&GNN&Reinforcement learning\\
\hline
DeepProLog\cite{manhaeve2018deepproblog}& \multirow{5}*{\tabincell{c}{{\em Learning\mbox{-}reasoning} \\ (Interaction)}} & \multirow{5}*{ \tabincell{c}{Interacting}} & \multirow{5}*{\tabincell{c}{First\mbox{-}order logic}} &\multirow{2}*{ \tabincell{c}{Agnostic}}& \multirow{4}*{\tabincell{c}{Complex reasoning}}\\
\cline{0-0}
ABL\cite{zhou2019abductive}& && &&\\
\cline{0-0}\cline{5-5}
GABL\cite{caiabductive}& & & &CNN&\\
\cline{0-0}\cline{5-5}
WS\mbox{-}NeSyL\cite{tian2022weakly}&  & &  & RNN&\\
\cline{1-1} \cline{6-6}\cline{5-5}
BPGR\cite{yu2022probabilistic}&  & & &CNN& Visual relationship detection\\ 
\hline
\end{tabular}}
\end{table*}

\section{Categorization and Frameworks}
\label{sec3}
In this section, we first introduce the theory and formal definition of neural-symbolic learning systems. We will then proceed to summarize the taxonomy of neural-symbolic methods and discuss the frameworks that underpin these taxonomies. In this survey, we use a color scheme to represent the different components of neural-symbolic learning systems: green represents symbolic systems, and blue represents neural systems within the frameworks.

\begin{figure}[htbp]
\centering
\includegraphics[width=0.8\linewidth]{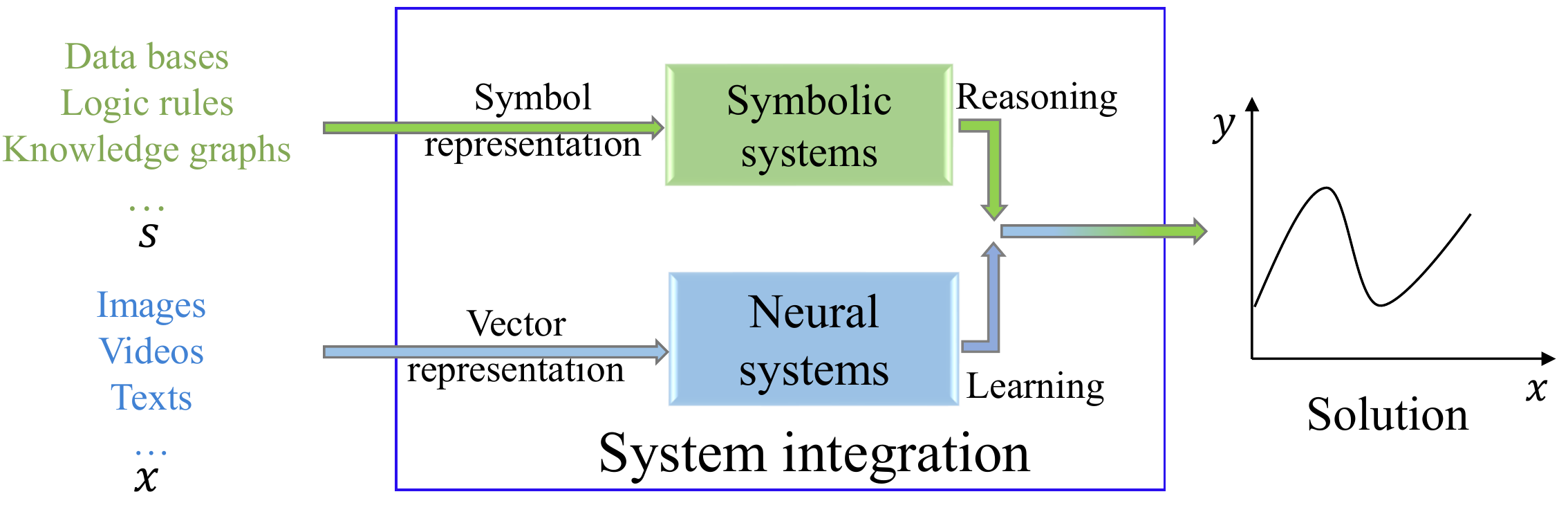}
\caption{Schematic diagram of two systems integration. Symbolic systems generally use reasoning technologies, such as logic programs and search algorithms, to obtain a solution based on domain knowledge, such as first\mbox{-}order logic, knowledge graphs, etc. The goal of neural systems is to learn a function from the training samples to predict a solution.}
\label{fig:3}
\end{figure}

Neural-symbolic learning systems are hybrid models that leverage the strengths of both neural systems and symbolic systems. To illustrate the relationships and characteristics between these systems, we provide a schematic diagram in Figure \ref{fig:3}. The green rectangle represents symbolic systems, which employ reasoning-based approaches to find solutions. Symbolic systems typically operate on structured data, such as logic rules, knowledge graphs, or time series data. Their fundamental unit of information processing is symbols. Through training, symbolic systems acquire the solution space of a search algorithm for a specific task and output higher-level reasoning results. On the other hand, the blue rectangle in the figure represents neural systems, which excel at learning-based approaches to approximate the ground truth. Neural systems usually operate on unstructured data, such as images, videos, or texts, and their primary information processing unit is a vector. Through training, neural systems learn a mapping function for a specific task and output lower-level learning results. The outer blue box represents neural-symbolic learning systems, which encompass the characteristics of both symbolic and neural systems (as summarized in Table \ref{1}). These systems combine the reasoning capabilities of symbolic systems with the learning capabilities of neural systems to achieve a comprehensive and integrated approach to problem-solving.

Based on the above description, the final objective of neural-symbolic learning systems is to find a function $F$ that can effectively map data $x$ and symbol $s$ (pre-defined or obtained via computation) to ground-truth $y$. The formal definition is as follows:
\begin{equation}\label{eq:define}
   \forall (x,y)\in D~ F(x,s) \rightarrow y.
\end{equation}

In this survey, the neural systems mainly refer to deep learning (or deep neural networks), while symbolic systems include symbolic knowledge \cite{tandon2018commonsense, davis2017logical, bosselut2019comet} and symbolic reasoning techniques \cite{darwiche2011sdd, safavian1991survey}, etc. The methodology of our classification is determined by the integration mode between neural systems and symbolic systems, which has three main integration methodologies. These three classifications are similar to \cite{kautz2022third} from essence but include it.

\subsection{Learning for reasoning}
\begin{figure}[htbp]
\centering
\includegraphics[width=0.8\linewidth]{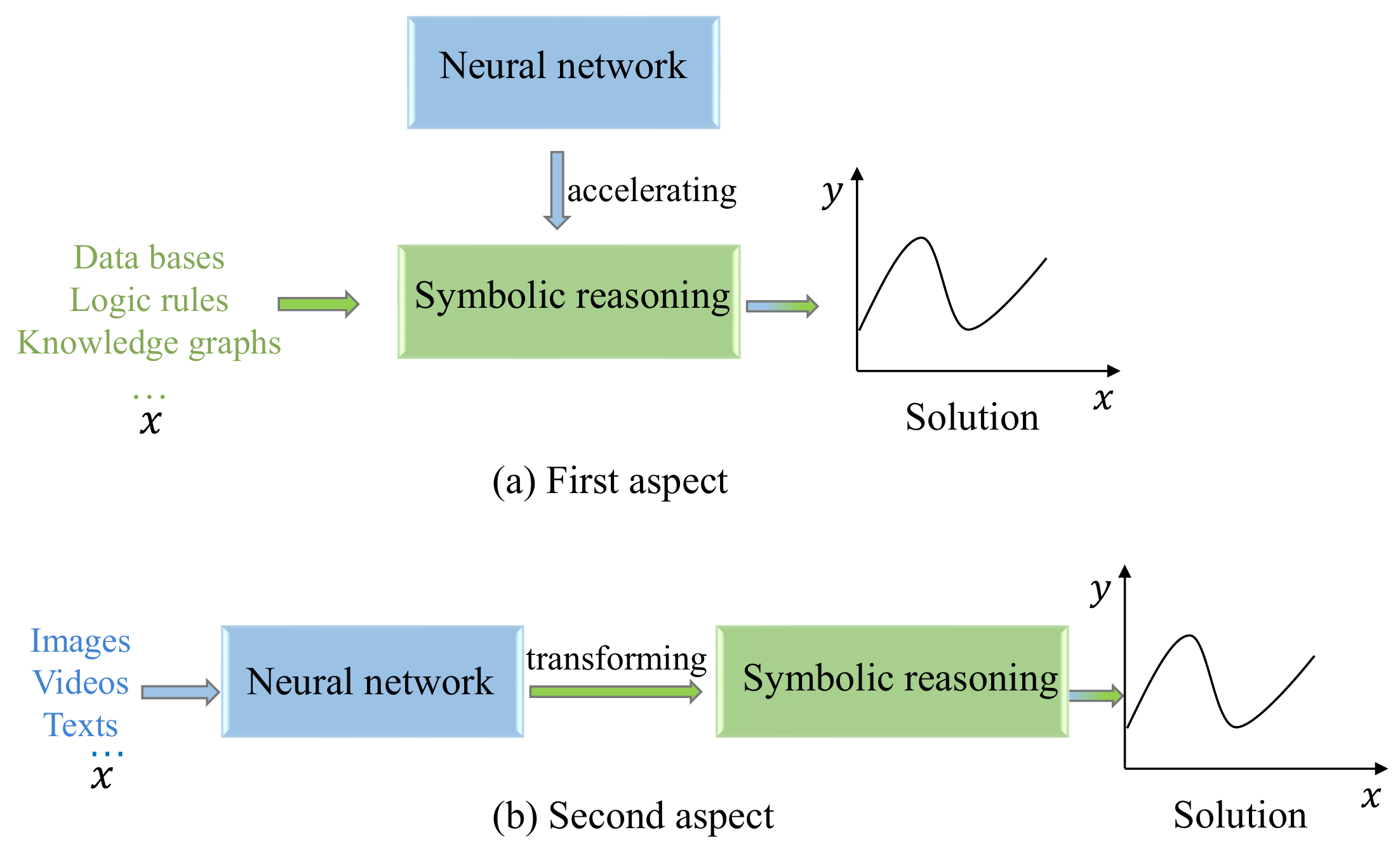}
\caption{Principle of {\em Learning for reasoning}. Its goal is to introduce neural networks to reasoning, a problem that it solves primarily through reasoning technologies.}
\label{learning for reasoning}
\end{figure}

In the first category, which we refer to as \textbf{\em learning for reasoning}, the goal is to leverage symbolic systems for reasoning while incorporating the advantages of neural networks to facilitate finding solutions \cite{qu2019probabilistic, zhang2020efficient, evans2018learning, yang2019learn, dos2019transforming, kalyan2018neural, ellis2018library, prates2019learning, nye2019learning, abboud2020learning, serafini2016logic, garcez2019neural, riegel2020logical,wang2019logic, neelakantan2015compositional, das2016chains, xiong2017deeppath, meilicke2020reinforced, das2017go, teru2020inductive, hudson2018compositional, hudson2019learning, garnelo2016towards, garcez2018towards}. In the context of \textit{learning for reasoning}, there are two main aspects to consider. The first aspect involves the use of neural networks to reduce the search space of symbolic systems, thereby accelerating computation \cite{qu2019probabilistic, zhang2020efficient, prates2019learning, nye2019learning, abboud2020learning}. This can be achieved by replacing traditional symbolic reasoning algorithms with neural networks. The neural network effectively reduces the search space, making the computation more efficient. The second aspect of \textit{learning for reasoning} is the abstraction or extraction of symbols from data using neural networks to facilitate symbolic reasoning  \cite{serafini2016logic, garcez2019neural, riegel2020logical}. In this case, neural networks serve as a means of acquiring knowledge for symbolic reasoning tasks. They learn to extract meaningful symbols from input data and use them for subsequent reasoning processes. The basic framework for \textit{learning for reasoning} is illustrated in Figure \ref{learning for reasoning}. As depicted in the figure, this type of model is characterized by a serialization process, where the neural network component and the symbolic reasoning component are connected sequentially. The neural network extracts relevant features or symbols from the input data, which are then used by the symbolic reasoning module to perform higher-level reasoning tasks.

\subsection{Reasoning for learning}
\begin{figure}[h]
\centering
\includegraphics[width=0.6\linewidth]{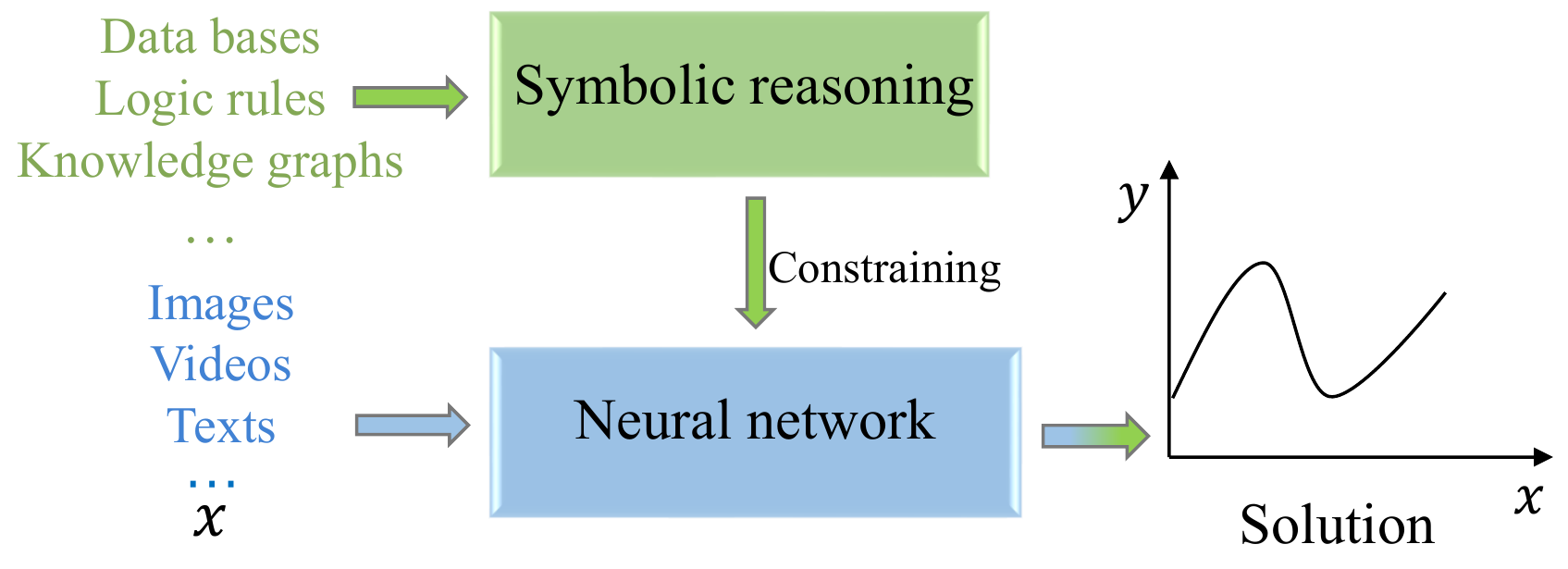}
\caption{Principle of {\em reasoning for learning}. It introduces symbolic knowledge to neural networks, and the main body relies on neural networks to get solutions.}
\label{fig:5}
\end{figure}

In the second category, referred to as \textbf{{\em reasoning for learning}}, symbolic systems are utilized to support the learning process of neural systems \cite{hu2016harnessing, diligenti2017semantic, xu2018semantic, xie2019embedding, luo2020context, donadello2017logic, minervini2017adversarial, chen2018neural,  bach2015hinge, li2019large, liu2019od, wang2018zero, kampffmeyer2019rethinking, chen2020knowledge,donadello2017logic,marszalek2007semantic,forestier2013coastal,zhu2014reasoning,nyga2014pr2,tran2008event,poon2009unsupervised, sun2020neural, oltramari2021generalizable, yang2018peorl,ji2022dual}. The underlying idea in \textit{reasoning for learning} is to leverage neural systems for machine learning tasks while incorporating symbolic knowledge into the training process to enhance performance and interpretability. Symbolic knowledge is typically encoded in a format suitable for neural networks and used to guide or constrain the learning process \cite{hu2016harnessing,diligenti2017semantic,xu2018semantic,xie2019embedding}. For instance, symbolic knowledge may be represented as a regularization term in the loss function of a specific task. This integration of symbolic knowledge helps improve the learning process and can lead to better generalization and interpretability of the neural models. The basic principle of the \textit{reasoning for learning} approach is illustrated in Figure \ref{fig:5}. This type of model is characterized by parallelization, where the neural system and symbolic system operate in parallel during the learning process. The neural network component learns from the data, while the symbolic system provides additional knowledge or constraints to guide the learning process.

\subsection{Learning\mbox{-}reasoning} 
\begin{figure}[htbp]
\centering
\includegraphics[width=0.8\linewidth]{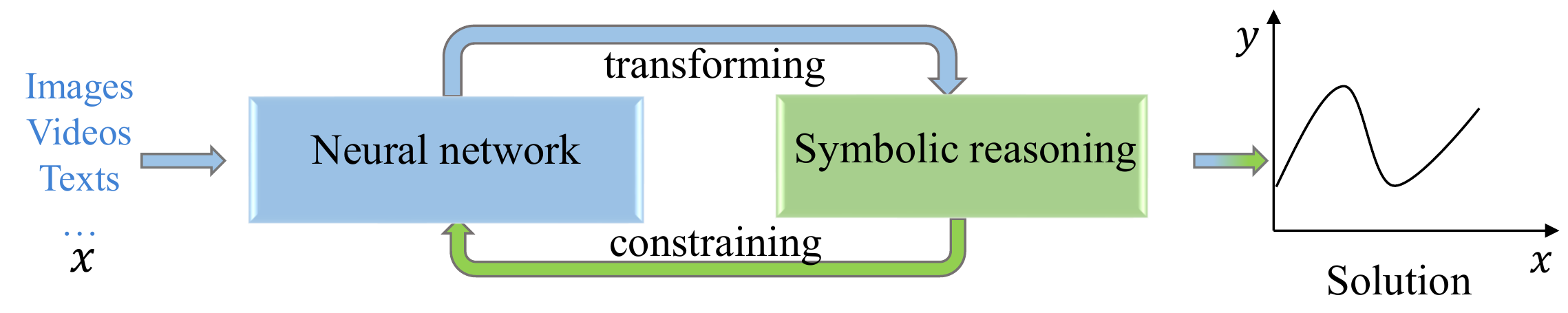}
\caption{Principle of {\em learning\mbox{-}reasoning}. It combines neural networks and reasoning technologies as an alternate process, and both together to output a solution.}
\label{fig:6}
\end{figure}

In the third category, referred to as \textbf{{\em learning-reasoning}}, the interaction between neural systems and symbolic systems is bidirectional, with both paradigms playing equal roles and working together in a mutually beneficial way \cite{manhaeve2018deepproblog, zhou2019abductive, caiabductive, tian2022weakly, yu2022probabilistic,gupta2019neural}. The goal of \textit{learning-reasoning} is to strike a balance between the involvement of neural systems and symbolic systems in the problem-solving process. In this approach, the output of the neural network becomes an input to the symbolic reasoning component, and the output of the symbolic reasoning becomes an input to the neural network. By allowing the neural systems and symbolic systems to exchange information and influence each other iteratively, this approach aims to leverage the strengths of both paradigms and enhance the overall problem-solving capability. For example, incorporating symbolic reasoning techniques like abduction enables the design of connections between deep neural networks and symbolic reasoning frameworks  \cite{manhaeve2019deepproblog,zhou2019abductive,yu2022probabilistic}. In this case, the neural network component generates hypotheses or predictions, which are then used by the symbolic reasoning component to perform logical reasoning or inference. The results from symbolic reasoning can subsequently be fed back to the neural network to refine and improve the predictions. The basic principle of \textit{learning-reasoning} is illustrated in Figure \ref{fig:6}, where the interaction between neural systems and symbolic systems occurs in an alternating fashion. This mode of combining both technologies allow for iterative learning and reasoning, enabling a deeper integration of neural and symbolic approaches. By embracing bidirectional interaction and iterative exchange of information between neural systems and symbolic systems, \textit{learning-reasoning} approaches aim to maximize the strengths of both paradigms and achieve enhanced problem-solving capabilities in various domains.

In summary, the above three taxonomies are  heterogeneous multi-module architectures in \cite{sun2013connectionist}. {\em Learning for reasoning} and {\em reasoning for learning} are loosely coupled while {\em learning-reasoning} is tightly coupled. According to the taxonomy of the neural-symbolic method presented in this paper, we summarize the existing main approaches from six dimensions in Table \ref{tab:2}: {\em representative works}, {\em taxonomies}, {\em methods}, {\em symbols}, {\em neural networks} and {\em applications}. In the Section \ref{sec4}, we will introduce the details of these approaches, discussing their methodologies, techniques, and characteristics. This will provide readers with a deeper understanding of how neural and symbolic systems are combined in various ways to tackle different problems.

\section{Methods of neural-symbolic learning systems}
\label{sec4}
This section introduces the methods used in neural-symbolic learning systems in three main categories. We aim to distill the representative ideas that provide evidence for the integration between neural networks and symbolic systems, identify the similarities and differences between different methods, and offer guidelines for researchers. The main characteristics of these representative methods are summarized in Table \ref{tab:5}.
\subsection{Learning for reasoning}
{\em Learning for reasoning} methods leverage neural networks to accelerate the search speed of symbolic reasoning, or to abstract unstructured data for symbolic reasoning. To accelerate symbolic reasoning, we first introduce approaches based on SRL \cite{koller2007introduction}, such as pLogicNet \cite{qu2019probabilistic} and ExpressGNN \cite{zhang2020efficient}. These approaches use neural networks to parameterize the posterior computation of the probabilistic graphical models, which accelerates the search process on solution space. Meanwhile, we investigate several methods based on ILP, such as NLIL \cite{yang2019learn}, in which NLIL automatically induces new logic rules from the data for model learning and reasoning. Next, for abstracting unstructured data, we introduce several representative approaches such as NS\mbox{-}CL\cite{mao2019neuro}. Specific details about the different models are presented below.

\begin{table*}[htbp]
\renewcommand{\arraystretch}{1.3}
\centering
\caption{Main characteristics of the selected methods.}
\label{tab:5}
\newcommand{\tabincell}[2]{\begin{tabular}{@{}#1@{}}#2\end{tabular}}
\resizebox{\linewidth}{!}{ 
\begin{tabular}{ccccl}
\hline
\textbf{Approaches}&\textbf{Inputs}&\textbf{Technology}&\textbf{Tools}&\textbf{Mechanism/Objective}\\
\hline
pLogicNet \cite{qu2019probabilistic}&x,s&SRL&MLN&learn a joint probability distribution\\
\hline
ExpressGNN \cite{zhang2020efficient}&x,s&SRL&MLN&learn a joint probability distribution\\
\hline
NLIL\cite{yang2019learn}&x&ILP&Transformer&use a Transform to learn rules based ILP\\
\hline
NS-CL\cite{mao2019neuro}&x&quasi-symbolic program& concept parser& \tabincell{l}{reason based on parsing symbols\\ for images and questions}  \\
\hline
HDNN\cite{hu2016harnessing}&x,s&regularization&t-norm&\tabincell{l}{learn a student network based on knowledge}\\
\hline
SBR\cite{diligenti2017semantic}&x,s&regularization&t-norm&\tabincell{l}{learn a model with logic knowledge \\as a constraint of the hypothesis space}\\
\hline
SL\cite{xu2018semantic}&x,s&regularization& arithmetic  circuits&\tabincell{l}{design a semantic loss to \\act as a regularization term}\\
\hline
LENSR\cite{xie2019embedding}&x,s&regularization&d-DNNF&\tabincell{l}{align distributions between deep learning\\ and propositional logic}  \\
\hline
CA\mbox{-}ZSL\cite{luo2020context}&x,s&regularization&GCN& learn a conditional random field\\
\hline
SEKB\mbox{-}ZSR\cite{wang2018zero}&x,s&knowledge transfer&GCN&\tabincell{l}{learn a deep learning model\\ with powerful generalization}\\
\hline
DGP\cite{kampffmeyer2019rethinking}&x,s&knowledge transfer&GCN& learn network embedding with semantics\\
\hline
KGTN \cite{chen2020knowledge}&x,s& knowledge transfer&GGNN&\tabincell{l}{transfer semantic knowledge into weights }\\
\hline
PROLONETS \cite{silva2021encoding}&x,s& knowledge transfer&decision tree&\tabincell{l}{transform knowledge into \\ neural network parameters }\\
\hline
DeepProLog \cite{manhaeve2018deepproblog}&x,s&ProbLog&SDD& \tabincell{l}{construct an interface between \\probLog program and the deep learning models}\\
\hline 
ABL\cite{zhou2019abductive}&x,s&abductive reasoning&SLD&\tabincell{l}{minimize inconsistency between \\pseudo\mbox{-}labels and symbolic knowledge} \\
\hline
WS\mbox{-}NeSyL\cite{tian2022weakly}&x,s&ProLog&SDD&\tabincell{l}{learn an encoder-decoder \\constrained by logic rules} \\
\hline
BPGR\cite{yu2022probabilistic}&x,s&SRL&MLN& \tabincell{l}{learn a model that fits both\\ the ground truth and FOL}\\
\hline
\end{tabular}}
\end{table*}

\subsubsection{Accelerating symbolic reasoning} 
In the case of Markov logic networks (MLNs), logic rules are encoded into an undirected graph, and graphical models' inference techniques are used to solve problems. However, performing computations on large-scale graphical models can be challenging. On the other hand, neural networks can easily scale to large datasets, but they cannot directly integrate logic knowledge. While several approximate inference methods have been proposed to address this issue, they often come with high computational costs  \cite{singla2005discriminative,mihalkova2007bottom,singla2006memory,poon2006sound,khot2011learning,bach2015hinge}. To overcome these challenges, researchers have introduced two models: Probabilistic Logic Neural Networks (\textbf{pLogicNet}) \cite{qu2019probabilistic} and \textbf{ExpressGNN} \cite{zhang2020efficient}. Both of these models aim to address the triplet completion problem in knowledge graphs by treating it as an inference problem involving hidden variables in a probability graph. They achieve this by combining variational expectation-maximization (EM) and neural networks to approximate the inference process. The basic process can be summarized as follows:

(1) Model Construction: Both pLogicNet and ExpressGNN start by constructing a joint probabilistic model that combines the logic rules encoded in the knowledge graph with the neural network model. This joint model captures the dependencies between observed and hidden variables. (2) Inference Approximation: To perform inference, pLogicNet and ExpressGNN utilize a variational EM approach. They define an approximate posterior distribution over the hidden variables and iteratively optimize the model parameters and approximate posterior distribution to maximize the joint probability. (2) Neural Network Learning: During the optimization process, the neural network parameters are updated based on the observed data, using techniques such as backpropagation and gradient descent. This allows the neural network to learn the underlying patterns and relationships in the data. (3) Inference and Prediction: After training, the models can be used for inference and prediction tasks. They can estimate the probabilities of unobserved variables, complete missing values, or make predictions based on the learned dependencies.

\begin{figure}[htbp]
\centering
\includegraphics[width=0.8\linewidth]{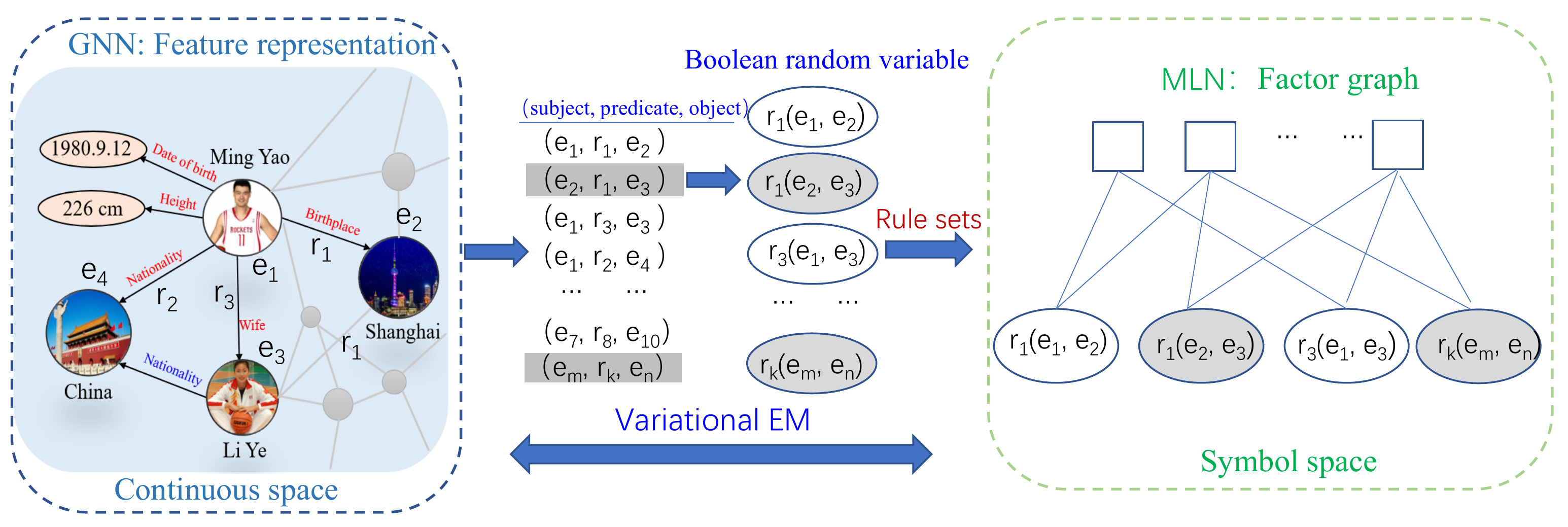}
\caption{The framework of the ExpressGNN model. The model contains continuous space and symbolic space, and the variational EM algorithm acts as a bridge that connects the continuous space and symbolic space. Note, in the knowledge graph, $e_k$ is the entity and $r_k$ is the relation.}
\label{fig:13}
\end{figure}

The main goal of pLogicNet and ExpressGNN is to combine the strengths of logic-based reasoning and neural networks in a unified framework. By leveraging neural networks' scalability and the expressiveness of logic rules, these models offer an solution for reasoning tasks in knowledge graphs.
ExpressGNN improves the inference network of pLogicNet by using a graph neural network (GNN) to replace the flattened embedding table and adding a tunable part to the entity embedding to alleviate the problem of isomorphic nodes having the same embedding. The specific framework of ExpressGNN is illustrated in Figure \ref{fig:13}.

Indeed, relying solely on manually constructed logic rules may not capture the complete knowledge present in the data. To address this limitation, researchers have explored approaches to automatically learn logic rules from data. Two notable methods in this regard are the extensions of Markov logic networks (MLNs) proposed by Marra et al. \cite{marra2019neural,marra2020relational} and differentiable inductive logic programming (ILP) models. Marra et al. extended MLNs by designing a general neural network architecture that can automatically learn the potential functions of MLNs from the original data. By training the neural network on labeled data, the model learns to capture the underlying patterns and relationships in the data, effectively learning logic rules that approximate the true structure of the domain. This approach enables the integration of neural networks and symbolic reasoning, allowing the model to learn logic rules directly from the data. Differentiable ILP is another approach that combines neural networks and logic to learn rules . It extends traditional ILP methods \cite{lavrac1994inductive} by introducing differentiable operations that enable the incorporation of neural networks into the ILP framework. This allows the model to learn logic rules by leveraging the expressive power of neural networks. Several differentiable ILP models have been proposed \cite{galarraga2015fast,evans2018learning,campero2018logical,rocktaschel2017end,payani2019inductive}, including $\partial$ILP \cite{evans2018learning}, which uses predefined templates to construct logic rules and applies forward reasoning for inference. $\partial$ILP is capable of learning effective logic rules even in the presence of noisy data, making it robust to imperfect input. 

The current methods for learning logic rules often face limitations in expressiveness and computational feasibility. Approaches such as expressing the chain rule as a Horn clause and controlling the search length, number of relationships, and entities have been employed to address these challenges \cite{das2016chains,yang2017differentiable,gardner2015efficient}. However, the limited expressive power of these complex logic rules can hinder their effectiveness. To overcome these limitations, Yang et al. introduced neural logic inductive learning (\textbf{NLIL}) \cite{yang2019learn}. NLIL is a differentiable ILP model that extends the multi-hop reasoning framework to address general ILP problems. It allows for the learning of complex logic rules, including tree and conjunctive rules, which offer greater expressiveness compared to traditional approaches. NLIL leverages neural networks to learn logic rules from data and provides explanations for patterns observed in the data.

It first converts a logical predicate into a predicate operation, and then transforms all the intermediate variables into predicate operation representations of the head and tail entities, and such the head and the tail variables can be represented by randomly initialized vectors in the concrete implementation, so as to get rid of data dependency; then such a predicate operation forms the atom of the logical paradigm, which greatly expands the expression capability of logical predicates such as from chains to trees. Next, the NLIL model further extends the expressive power of the generated logical paradigm by combining atoms using logical connectives (and, or, not). Indeed, the NLIL model uses a hierarchical transformer model \footnote{Transformer is a tool that learns the relation between sequence data. Here, learning rules is a process of sequence calculation.} to efficiently compute the intermediate parameters to be learned, including the vectors of logical predicates and the corresponding parameters of the attention mechanism (the weighted summation).


In NLIL, logic rules are grounded through matrix multiplication. For example, consider the logic rule $Friends(x,y)\Rightarrow Smokes(x)$, where constants $C=\{A, B\}$ are one-hot vectors, such as $V_A$ and $V_B$, and predicates $Friends$ and $Smokes$ are mapped to matrices such as $M_{Friend}(A, B)$ is a score of $A$ and $B$ that are related by $Friend$. The score of grounding is $V_B=V_{A}M_{Friend}$.


\subsubsection{Abstracting unstructured data into symbols}
Mao et al. \cite{mao2019neuro} proposed the Neuro-Symbolic Concept Learner (NS-CL), which uses neural symbolic reasoning as a bridge to jointly learn visual concepts, words, and the semantic parsing of sentences without the need for the explicit supervision of any of them. NS\mbox{-}CL builds an object-based scene representation and translates sentences into symbolic programs.

NS\mbox{-}CL is designed to solve tasks in visual question answering (VQA). This model includes three modules: a {\em visual perception} module, {\em semantic parsing} module, and {\em symbolic reasoning} module. The {\em visual perception } module extracts object-based symbolic representation for a scene in an image. The {\em  semantic parsing} module transforms a question into an executable program. Finally, the {\em symbolic reasoning} module applies a quasi-symbolic program executor to infer the answer based on the representation and an executable program.

\begin{figure}[htbp]
\centering
\includegraphics[width=0.8\linewidth]{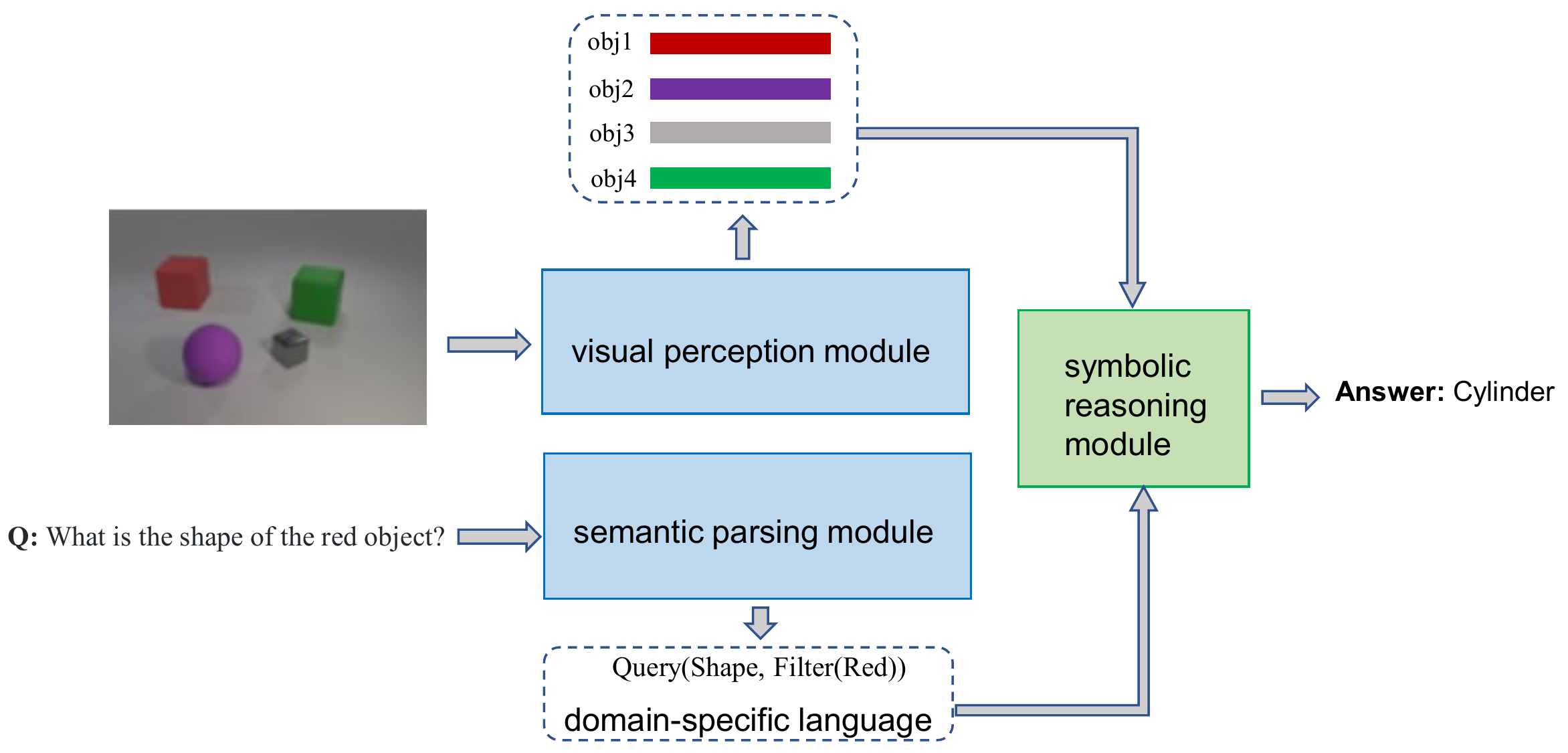}
\caption{The framework of the NS-CL. The perception module begins by parsing visual scenes
into object-based deep representations, while the semantic parser parse sentences into executable programs. A symbolic reasoning process bridges two modules.}
\label{fig:NS-CL}
\end{figure}

An example is shown in Figure \ref{fig:NS-CL}. Given the input image and question, the {\em visual perception} module uses a pre-trained mask R\mbox{-}CNN
\cite{dollar2017mask} to generate object features for all objects, then applies a similarity-based metric to classify objects. The {\em semantic parsing} module translates a natural language question into an executable program with a hierarchy of primitive operations, represented in a domain-specific language (DSL) designed for VQA. The {\em symbolic reasoning} module is a collection of deterministic functional modules designed to realize all logic operations specified in the DSL. It takes object concepts and a program as input to derive the answer. The optimization objective of NS\mbox{-}CL is composed of both a {\em visual perception} module and a {\em semantic parsing} module.

\textbf{Conclusion:} Based on the discussion of related works in the category of \textit{learning for reasoning}, we can summarize two principles from the perspective of neural networks: (1) \textbf{Accelerator}. Symbolic reasoning often involves searching for solutions in a solution space, which can be computationally expensive. In \textit{learning for reasoning} approaches, neural networks are used as accelerators to speed up the search process. By leveraging the learning capabilities of neural networks, the search space can be reduced or optimized, leading to faster and more efficient reasoning. Techniques such as reinforcement learning can be employed to guide the search process and improve the overall performance of symbolic systems. (2) \textbf{Transformer}. Symbolic systems traditionally operate on structured data represented by symbols. However, neural networks excel at processing unstructured data such as images, texts, and videos. In \textit{learning for reasoning} approaches, neural networks play a crucial role in transforming unstructured data into symbolic representations that can be processed by symbolic reasoning systems. By abstracting and extracting relevant features from the unstructured data, neural networks provide the symbolic systems with meaningful inputs for reasoning and decision-making.


\subsection{Reasoning for learning}
{\em Reasoning for learning} methods can be broadly divided into regularization models and knowledge transfer models. \textbf{Regularization models}: Regularization models integrate symbolic knowledge into the training process by adding regular terms to the objective function of the model. These regular terms serve as constraints or penalties that encourage the model to adhere to the symbolic knowledge during training. By incorporating prior knowledge, regularization models aim to improve the performance and interpretability of the trained model. Different regularization models may employ different strategies for modeling symbolic knowledge as regular terms, such as semantic embeddings, logical rules, or semantic constraints. Examples of regularization models include \cite{hu2016harnessing,diligenti2017semantic,xu2018semantic,xie2019embedding,luo2020context,donadello2017logic,minervini2017adversarial,chen2018neural,li2019large,liu2019od}. \textbf{Knowledge transfer models}: Knowledge transfer models establish connections between different domains or spaces, such as the visual space and the semantic space, and transfer symbolic knowledge from one domain to another to support the learning process. These models leverage existing symbolic knowledge or semantic information to guide the learning of models in a different domain. For example, semantic knowledge can be transferred from the semantic space to the visual space to aid in tasks such as zero-shot learning or semantic segmentation. Knowledge transfer models enable the utilization of symbolic knowledge in a different context to improve the performance and generalization of the model. Examples of knowledge transfer models include  \cite{wang2018zero,kampffmeyer2019rethinking,chen2020knowledge}.

\subsubsection{Regularization models}
Indeed, Hu et al. \cite{hu2016harnessing} introduced the \textbf{HDNN} framework, which leverages the concept of knowledge distillation to harness the power of logic rules in training deep neural networks. The framework consists of two components: a teacher network and a student network. In HDNN, the teacher network encodes logic rules and guides the student network during training. That is to say, the teacher network learns information from the labeled data and logic rules (unlabeled data) and teaches the student network by the loss function. Based on the above process, the structured information encoded by logic rules can constrain the learning of the student network. 

Specifically, the {\em rule knowledge distillation} module includes the student network $p_\theta(y\vert x)$ and the teacher network $q(y\vert x)$. The model needs to meet the following conditions: 1) The probability distribution $p_\theta(y\vert x)$ of the student network should be as close as possible to the probability distribution $q(y\vert x)$ of the teacher network; 2) The teacher network should obey the logic rules to the greatest extent possible. The formal representation of the objective and two conditions is shown in Eq. (\ref{eq:14}) and Eq. (\ref{eq:15}). 

\begin{equation}\label{eq:14}
    \theta^{(t+1)}={\arg\min}_{\theta \in \Theta} 1/N\sum_{n=1}^N(1-\pi) l(y_n, \sigma_\theta(x_n))
    +\pi l({s_n}^{(t)}, \sigma_\theta(x_n)),
\end{equation}
\begin{equation}\label{eq:15}
   \begin{cases}
    \min_{q,\xi\geq 0}KL(q(Y\vert X)\Vert p_\theta(Y\vert X))+C\sum_{l,gl}\xi_{l,gl}\\
    \lambda_l(1-E_q[r_{l,gl}(X,Y)])\leq \xi_{l,gl}\\
    gl=1,...,G_l, l=1,...,L
    \end{cases}
\end{equation}

In Eq. (\ref{eq:14}), $\pi$ is the limitation parameter used to calibrate the relative importance of the two objectives; $x_n$ represents the training data, while $y_n$ the label of the training data; $l$ denotes the loss function selected according to specific applications (e.g., the cross-entropy loss for classification); $s_n^{(t)}$ is the soft prediction vector of $q(y\vert x)$ on $x_n$ at iteration $t$; $\sigma_\theta(x)$ represents the output of $p_\theta(y\vert x)$; the first term is the student network, and the second term is the teacher network. In Eq. (\ref{eq:15}), $\xi_{l,gl}\geq 0$ is the slack variable for the respective logic constraint; $C$ is the regularization parameter; $l$ is the index of the rule; $gl$ is the index of the ground rule; $\lambda_l$ is the weight of the rule.


Different from the knowledge distillation framework, certain approaches incorporate logical knowledge as a constraint within the hypothesis space. These methods involve encoding a logic formula, either propositional or first-order, into a real-valued function that serves as a regularization term for the neural model. An example of such an approach is semantic-based regularization (SBR) proposed by Diligenti et al. \cite{diligenti2017semantic}. SBR combines the strengths of classic machine learning, with its ability to learn continuous feature representations, and symbolic reasoning techniques, with their advanced semantic knowledge reasoning capabilities. SBR is applied to address various problems, including multi-task optimization and classification. Following the classical penalty approach for constrained optimization, constraint satisfaction can be enforced by adding a term that penalizes the violation of these constraints into the loss of the model.   

Building upon the idea of semantic-based regularization (SBR), Xu et al. \cite{xu2018semantic} introduced a novel approach called semantic loss (\textbf{SL}). SL combines the power of propositional logic reasoning with deep learning architectures by incorporating the output of the neural network into the loss function as a constraint for the learnable network. This enables the neural network to leverage the reasoning capabilities of propositional logic to improve its learning ability. In contrast to SBR, SL takes a different approach to incorporating logic rules into the loss function. It encodes the logic rules using an arithmetic circuit, specifically a Sentential Decision Diagram (SDD) \cite{darwiche2011sdd}, which allows for the evaluation of the model. This encoding serves as an additional regularization term that can be directly integrated into an existing loss function. The formulation of SL is provided in Equation (\ref{eq:SL}).
\begin{equation}\label{eq:SL}
    L^s(\alpha, p) \propto -\log \sum_{x\models \alpha}\prod_{i:x\models X_i}p_i\prod_{i:x\models \neg X_i}(1-p_i),
\end{equation}
where $p$ is a vector of probabilities from the prediction of the neural network, $\alpha$ is a propositional logic, $x$ is the instantiation of $X_i$, and $x\models \alpha$ is a state $x$ that satisfies a sentence $\alpha$.

Notably, the aforementioned methods do not employ explicit knowledge representation techniques, leading to an unclear computational process for symbolic knowledge. To address this issue, some researchers have chosen to utilize tools that can model symbols, such as d-DNNF (as mentioned in Section \ref{sec2}). Xie et al. \cite{xie2019embedding} integrated propositional logic into a relationship detection model and proposed a logic embedding network with semantic regularization (\textbf{LENSR}) to enhance the relationship detection capabilities of deep models. The process of LENSR can be summarized as follows: 1) The visual relationship detection model predicts the probability distribution of the relation predicate for each image; 2) The prior propositional logic formula related to the sample image is expressed as a directed acyclic graph by d-DNNF, after which GNN is used to learn its probability distribution; 3) An objective function is designed that aligns the above two distributions. 

\begin{figure}[htbp]
\centering
\includegraphics[width=0.6\linewidth]{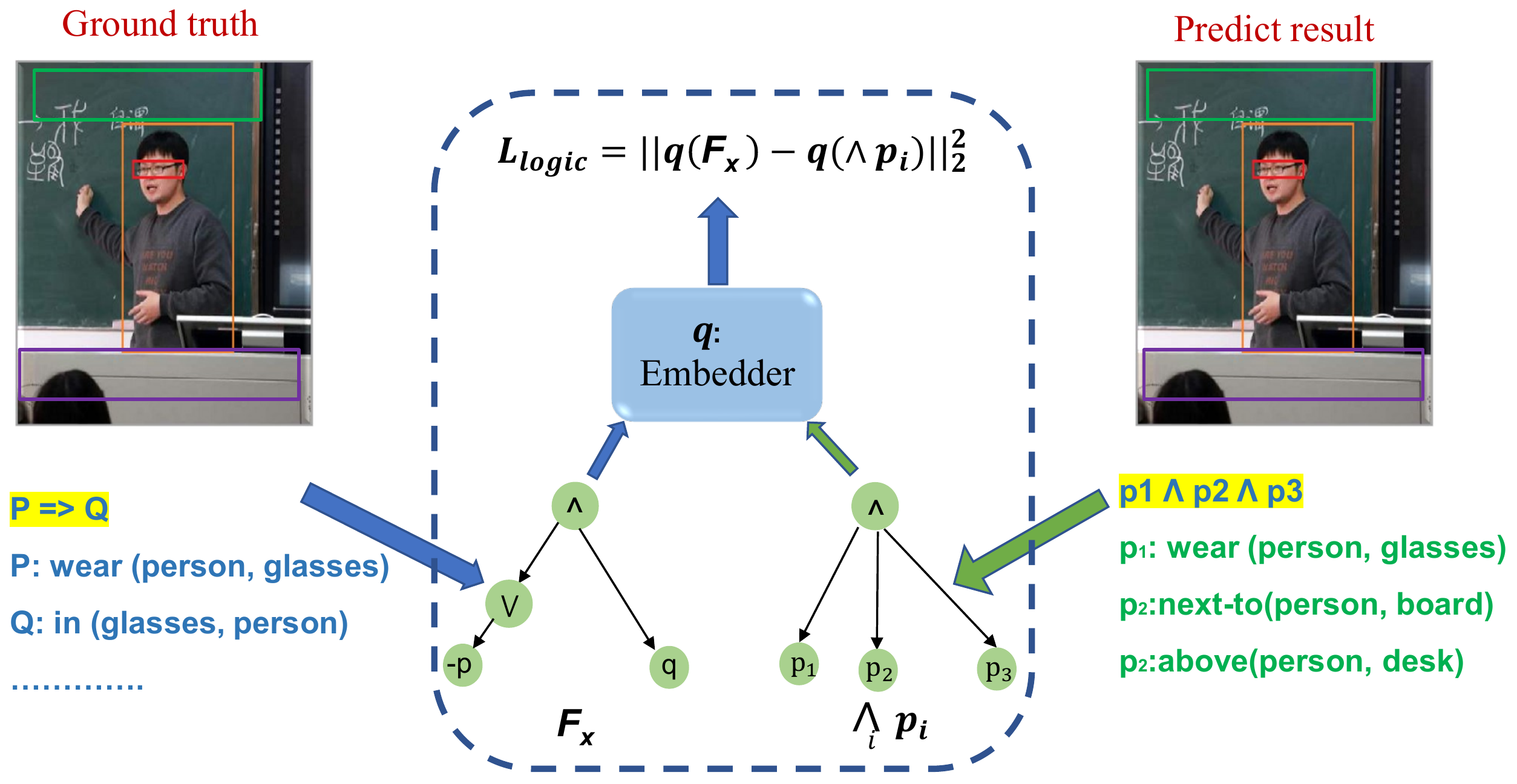}
\caption{The framework of the LENSR. The GCN-based embedder $q$ projects a logic graph to the vector space, satisfying the requirement that the distribution of the projected result is as close to the distribution of the real label as possible. We use this embedding space to form logic losses that regularize deep neural networks for a target task.}
\label{fig:16}
\end{figure}

Figure \ref{fig:16} presents a schematic diagram of LENSR, which uses a propositional logic of the form  $P\Rightarrow Q$. In this example, the predicate $P$ represents $wear(person, glasses)$, and the predicate $Q$ represents $in(glasses, person)$. The ground truth of the input image and the corresponding propositional logic (prior knowledge) are on the left. The directed acyclic graph of propositional logic d-DNNF is then sent to the Embedder $q$ (Embedder is a graph neural network (GNN) \cite{kipf2016semi} that learns the vector representation) to obtain $q(F_x)$, which is the embedding of the propositional logic knowledge. On the right is the relation label predicted by the detection network. The predicted labels are then combined into a conjunctive normal form $h(x)= \wedge p_i$ to construct a directed acyclic graph d-DNNF, which is sent to the Embedder $q$ to obtain  $q(h(x))$, the embedding of the predicted propositional logic. The optimization goal of LENSR is shown in Eq. (\ref{eq:16}). Here, $L_{task}$ represents the loss of a specific task, $\lambda$ is a hyperparameter that acts as a balance factor, and $L_{logic}$ is the loss of propositional logic (that is, the distance between vector $q(Fx)$ and vector $q(h(x))$).
\begin{equation}\label{eq:16}
    L=L_{task}+\lambda L_{logic} ,
\end{equation}
\begin{equation}\label{eq:17}
   L_{logic}=\Vert q(f)-q(\wedge p_i) \Vert_2 .
\end{equation}

However, LENSR models a local dependency graph for each logic formula and only captures local knowledge information, which may result in the poor expressive ability of the model. To solve this problem, the researcher started to model a global dependency graph for all logic formulas, which can improve the expressive ability of the model and effectively capture uncertainty \cite{yu2022probabilistic}.

\begin{figure}[htbp]
\centering
\includegraphics[width=0.8\linewidth]{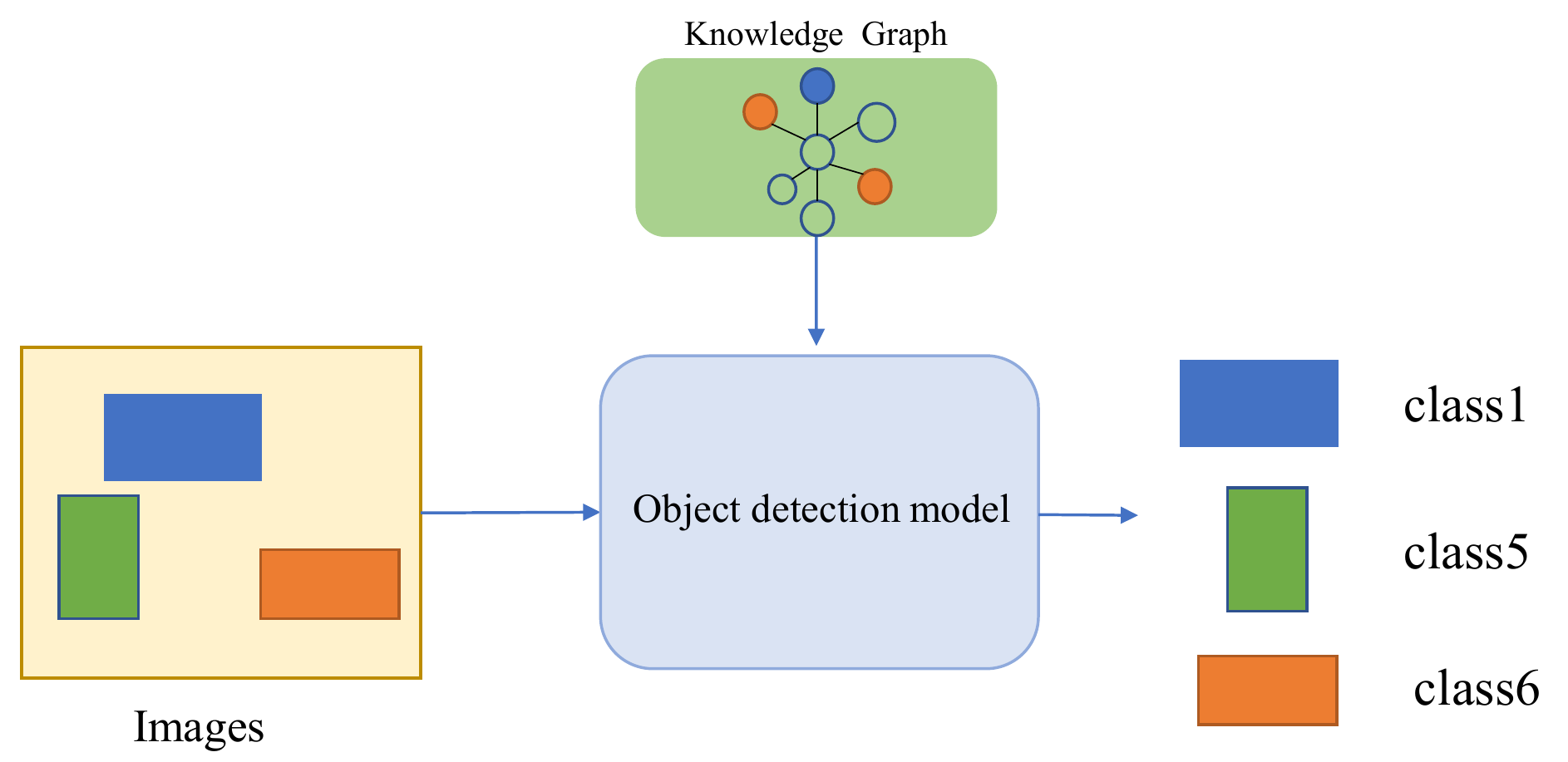}
\caption{The framework of the CA\mbox{-}ZSL. The features of individual objects and pairwise features are extracted from the image and input into an instance-level zero\mbox{-}shot inference module and a relationship inference module respectively. In combination with the knowledge graph, the unary potential function and binary potential function of CRF are generated respectively to predict the labels of objects.}
\label{fig:17}
\end{figure}

The above methods utilize logic rules as prior knowledge. In contrast, Luo et al. \cite{luo2020context} focused on knowledge graphs and proposed a context-aware zero-shot recognition method (\textbf{CA-ZSL}) to address the zero-shot detection problem. CA-ZSL constructs a model based on deep learning and conditional random fields (CRF) and leverages knowledge graphs, which represent the semantic relationships between classes, to assist in identifying objects from unseen classes. The framework of CA-ZSL is depicted in Figure \ref{fig:17}. In this framework, individual and pairwise features are extracted from the image. The instance-level zero-shot inference module utilizes individual features to generate a unary potential function, while the relationship inference module employs pairwise features and knowledge graphs to generate a binary potential function. Finally, based on the conditional random field constructed by these two potential functions, the label of the unseen objects is predicted. 

CA-ZSL incorporates a knowledge graph, which includes the GCN-encoded embedding, into the computation of the binary potential function within the conditional random field, thereby facilitating the learning process of the model. The objective of optimization is to maximize the joint probability distribution of the conditional random field, as depicted in Equation (\ref{eq:18}). In the equation, $\theta$ denotes the unary potential function, $\psi$ represents the binary potential function, $c_i$ represents the class, $B_i$ represents the object region in the image, $\gamma$ is the balance factor, and $N$ denotes the number of objects.
\begin{equation}\label{eq:18}
\footnotesize
    P=(c_1...c_N\vert B_1...B_N)
    \propto exp(\sum_i\theta(c_i\vert B_i)+\gamma \sum_{i\neq j}\psi(c_i,c_j\vert B_i,B_j)).
\end{equation}

\subsubsection{Knowledge transfer models}
Knowledge transfer integrates knowledge graphs (pre-defined) that represent semantic information into deep learning models, which compensates for the lack of available data through the transfer of semantic knowledge. In this survey, we mainly introduce representative approaches in image classification \cite{wang2020generalizing,chen2020knowledge} such as  zero\mbox{-}shot learning \cite{wang2018zero,kampffmeyer2019rethinking}, few-shot learning \cite{zhong2019decadal}  \cite{wang2020generalizing,chen2020knowledge} and reinforcement learning \cite{silva2021encoding} for knowledge transfer.

In the context of visual tasks, researchers have proposed several zero-shot recognition models that leverage semantic representation and knowledge graphs to compensate for the limitations of visual data. Two notable models in this regard are \textbf{SEKB-ZSL} (Zero-shot Recognition via Semantic Embeddings) \cite{wang2018zero} and \textbf{DGP} (Dense Graph Propagation Module) \cite{kampffmeyer2019rethinking}. These models employ the semantic classifier weights derived from the knowledge graph, which encompasses both seen and unseen classes, to guide or supervise the learning of visual classifier weights, thus facilitating knowledge transfer. The underlying principles can be summarized as follows: In the visual space, SEKB-ZSL employs Convolutional Neural Networks (CNN) to extract visual features from images and learns a visual classifier for the seen classes. In the semantic space, Graph Convolutional Networks (GCN) are utilized to learn node features within the knowledge graph and derive a semantic classifier for class labels. Subsequently, the weights of the class semantic classifier are employed to supervise the learning process of visual classifier weights, enabling the transfer of semantic knowledge to novel classes.

\begin{figure}[htbp]
\centering
\includegraphics[width=0.8\linewidth]{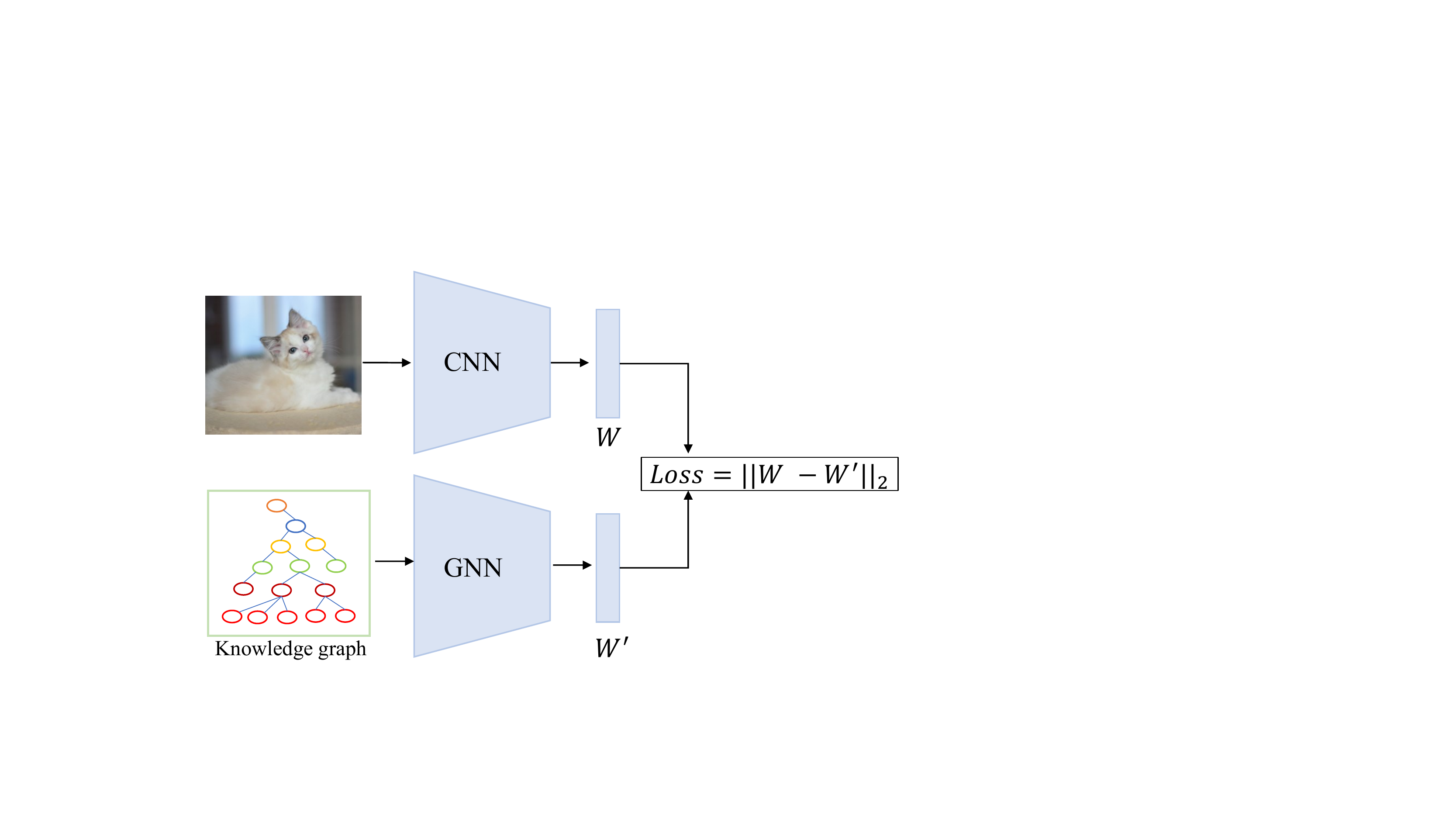}
\caption{The framework of the DGP. DGP is trained to predict the classifier weights $W$ for each node/class in a graph. The weights for the training classes are extracted from the final layer of a pre-trained ResNet. The graph is constructed from a knowledge graph, and each node is represented by a vector that encodes semantic class information (the word embedding class in this paper). The network consists of two phases: a descendant phase (where each node receives knowledge from its descendants) and an ancestor phase (where it receives knowledge from its ancestors).}
\label{fig:19}
\end{figure}

DGP introduces improvements over SEKB-ZSL. Firstly, it addresses the over-smoothing issue in GCN by reducing the number of graph convolution layers from six to two. Additionally, the attention mechanism is employed to calculate connection weights between nodes in the knowledge graph, thereby enhancing the node connectivity. The principle of DGP is depicted in Figure \ref{fig:19}. The entire model is trained end-to-end, and the loss function is defined as the similarity between the weights of the two module classifiers, as shown in Eq. (\ref{eq:20}). In the equation, $M$ represents the number of classes, $P$ represents the dimension of the weight vectors, $W_{i,j}$ denotes the weights of the visual classifier, and ${W'}_{i,j}$ denotes the weights of the semantic classifier.


\begin{equation}\label{eq:20}
    L=1/2M\sum_{i=1}^M\sum_{j=1}^P{(W_{i,j}-{W'}_{i,j})}^2.
\end{equation}

\begin{figure}[htbp]
\centering
\includegraphics[width=0.8\linewidth]{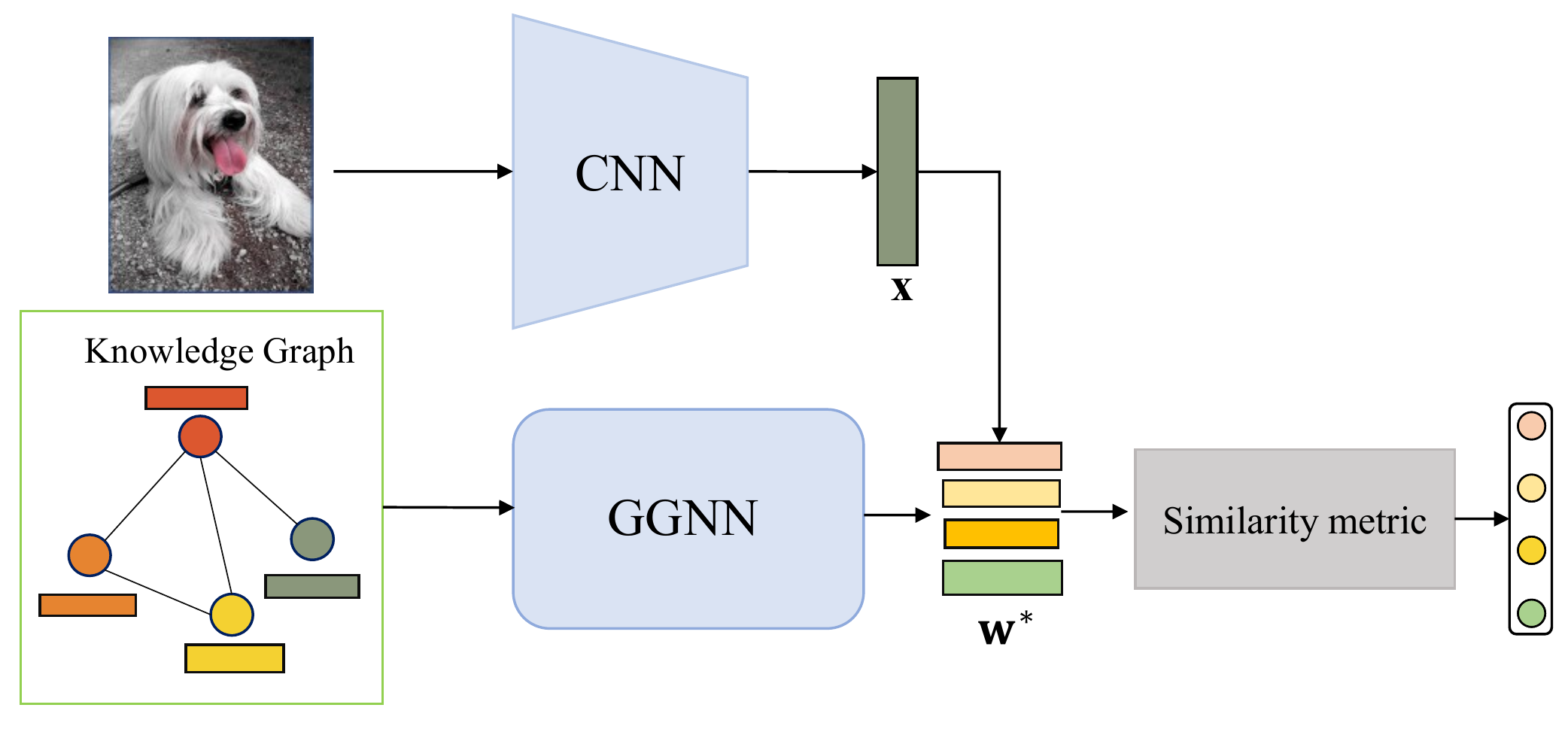}
\caption{The framework of the KGTN model. It incorporates the prior knowledge of category correlation and makes use of the interaction between category classifier weights, facilitating better learning of the classifier weights of unknown categories. }
\label{fig:20}
\end{figure}

Transferring correlation information between classes can be beneficial for learning new concepts. DGP demonstrates the importance of aligning the semantic classifier with the feature classifier. Building upon this idea, Chen et al. \cite{chen2020knowledge} proposed the Knowledge Graph Transfer Network (\textbf{KGTN}) to address the few-shot classification problem. In KGTN, knowledge graphs are utilized to capture and model the correlations between seen and unseen classes. These knowledge graphs serve as a means to transfer knowledge and facilitate the learning process. The overall architecture of KGTN is depicted in Figure \ref{fig:20}.

Specifically, KGTN comprises three main parts: the {\em feature extraction} module, {\em knowledge graph transfer} module, and {\em prediction} module. The {\em feature extraction} module uses CNN to extract the feature vector of images. The {\em knowledge graph transfer} module uses a gated graph neural network (GGNN) to learn the knowledge graph node embedding. After $T$ iterations, the {\em knowledge graph transfer} module obtains the final weight $w^*$, which has captured the correlation between the seen and the unseen classes. The {\em prediction} module calculates the similarity between the weight $w^*$ and the image feature to predict the probability distribution of the label.

\begin{figure}[htbp]
\centering
\includegraphics[width=0.8\linewidth]{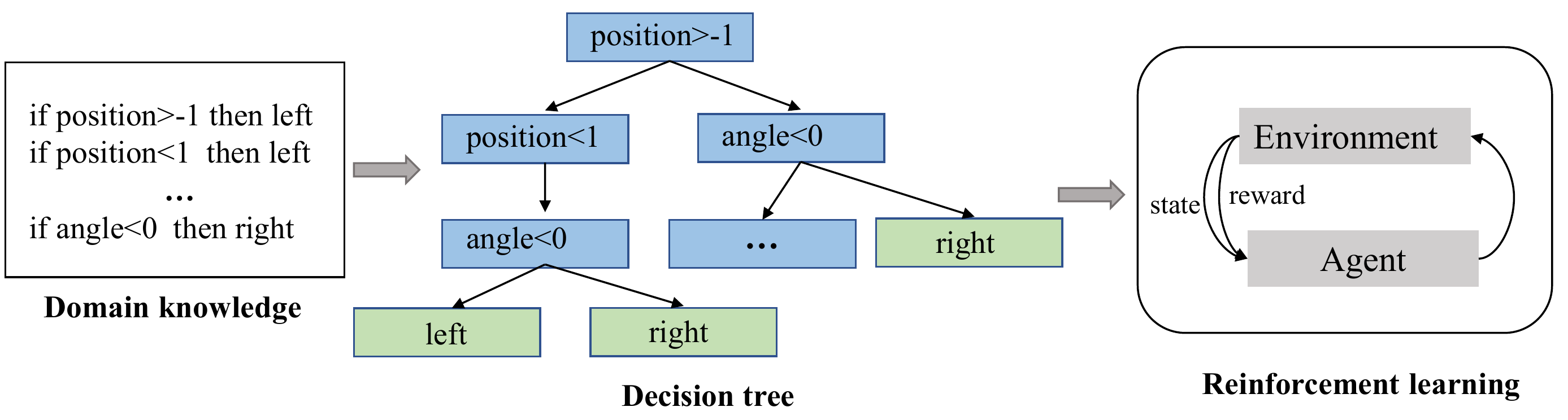}
\caption{The framework of the PROLONETS. Domain knowledge is constructed into a decision tree that is then used to directly initialize a PROLONET’s architecture and parameters; here, leaf nodes are actions and other nodes are policies in reinforcement learning. The PROLONET can then begin reinforcement learning in the given domain, outgrowing its original specification.}
\label{fig:reinforcement}
\end{figure}

In contrast to static domain knowledge, Silva et al. \cite{silva2021encoding} introduced Propositional Logic Nets (\textbf{PROLONETS}), which directly encode domain knowledge as a collection of propositional rules within a neural network. PROLONETS not only incorporates domain knowledge into the model but also allows for the refinement of domain knowledge based on the trained neural network. The framework of PROLONETS is illustrated in Figure \ref{fig:reinforcement}. This approach enables the neural network to leverage domain-specific information and improve its learning and reasoning capabilities.

PROLONETS aids in ``warm starting" the learning process in deep reinforcement learning. The first step involves knowledge representation, where policies and actions express domain knowledge in the form of propositional rules. These rules are then encoded into a decision tree structure. The second step is neural network initialization, wherein the nodes of the decision tree are directly transformed into neural network weights. This allows the agent to immediately commence learning effective strategies in reinforcement learning. The final step is training, during which the initialized network interacts with the environment, collecting data that is subsequently used to update parameters and rectify domain knowledge.

Let us consider the cart pole as an example. The state space of a cart pole is a four-dimensional vector: cart position, cart velocity, pole angle, and pole velocity. The action space is a two-dimensional vector (left, right). Domain knowledge can be expressed as "{\em if the cart’s position is right of center, move left; otherwise, move right.}". The decision nodes of the tree become linear layers, leaves become action
weights, and the final output is a sum of the leaves weighted by path probabilities. Therefore, if $position \textgreater -1$, the weight of the neural network is $w=\{1,0,0,0\}$, and the bias is $b=-1$.

\textbf{Conclusion:} Based on the above text, we can summarize the following key factors in {\em reasoning for learning}. (1) \emph{Knowledge representation}. Symbolic knowledge is a kind of discrete representation. To achieve combination between symbolic knowledge (discrete representation) and neural network (continuous representation), most methods usually convert the symbolic knowledge into an intermediate representation, such as a graph, tree, etc. Moreover, another approaches use fuzzy logic (such as t-norm) to assign soft truth degrees in the continuous set [0, 1]. In summary, it may very well be among the essential representations grounded in the
environment that form the foundation of a much larger representational edifice that is needed for human-like general intelligence \cite{honavar1995symbolic}. (2) \emph{Combining approaches}. One type of approach involves taking symbolic knowledge as a regularization term in the loss functions of the neural networks. The others involve encoding symbolic knowledge into the structure of the neural networks as an initiative to improve their performance. It is worth noting that logic rules usually are added as constraints to loss functions, while knowledge graphs often enhance neural networks with information about relations between instances.
\subsection{Learning\mbox{-}reasoning}
In {\em learning\mbox{-}reasoning} approaches, learning and reasoning do not work in isolation but instead closely interact. This is a development trend of neural-symbolic learning systems \cite{manhaeve2018deepproblog,zhou2019abductive,caiabductive,tian2022weakly,yu2022probabilistic}.

Based on ProbLog \cite{de2007problog}, Robin et al. \cite{manhaeve2018deepproblog} introduced neural facts and neural annotated disjunction (neural AD) to propose a model that seamlessly integrates probability, logic, and deep learning, known as \textbf{DeepProbLog}. DeepProbLog is a pioneering framework that combines a generic deep neural network with probabilistic logic in a unique manner. It offers the advantage of enhanced expressive capability and enables end-to-end training for neural networks and logical reasoning in a unified framework.

\begin{figure}[htbp]
\centering
\includegraphics[width=0.6\linewidth]{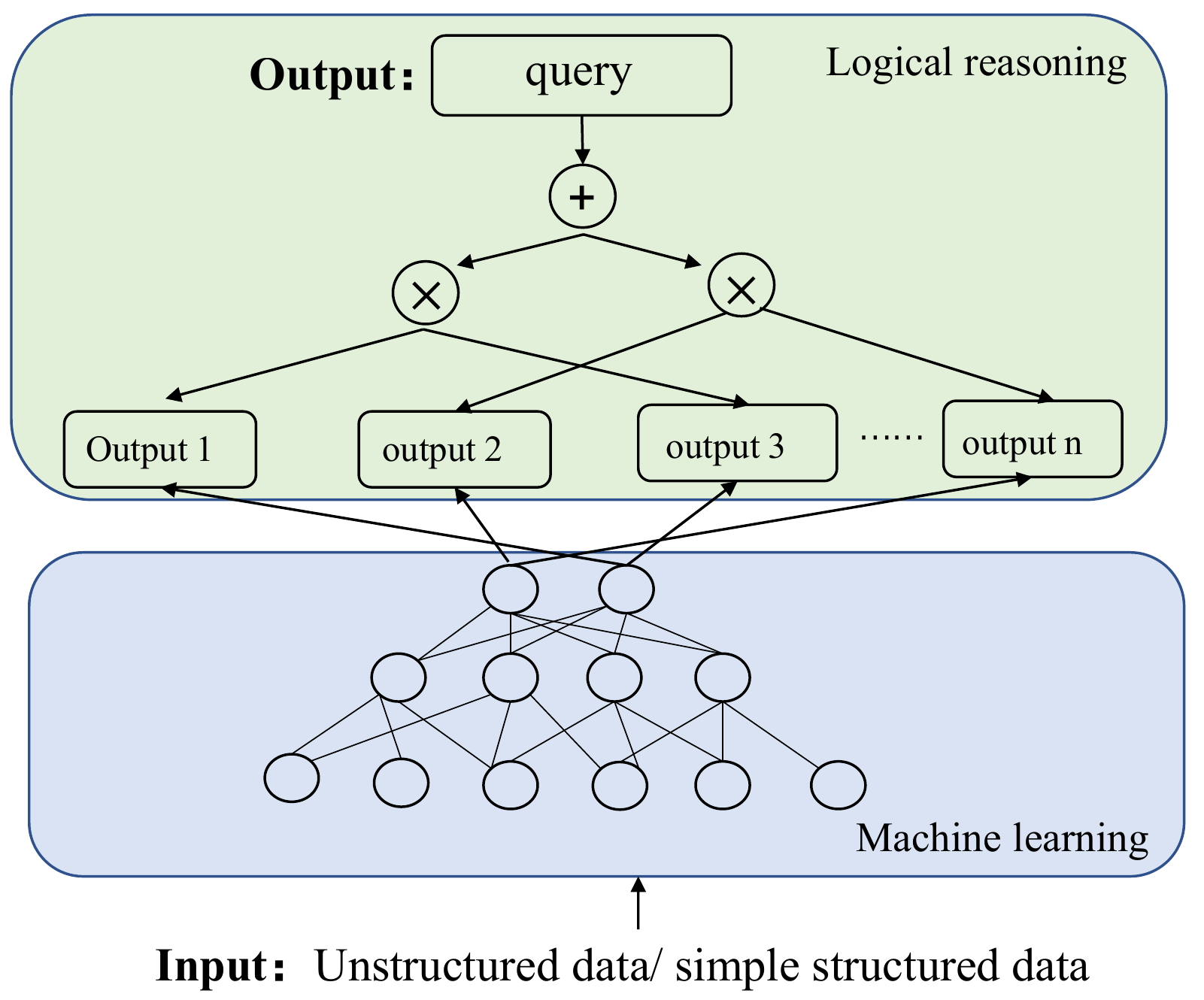}
\caption{The framework of DeepProLog. Machine learning is responsible for mapping the input (unstructured data or simple structured data) to the distribution of categories (if there are n categories, the output distribution is 1×n). Logical reasoning is a complex problem described by ProbLog, which is constructed as an arithmetic circuit to solve the complex problem. Here, the root node is the query, while the leaf nodes are neural predicates and other (non-neural network output) probabilistic facts.}
\label{fig:deepprolog}
\end{figure}

DeepProbLog is a probabilistic programming language that integrates deep learning through the use of "neural predicates". These neural predicates serve as an interface between neural networks and symbolic reasoning. In DeepProbLog, an image, for instance, is processed by a neural network, which outputs the distribution of each class in the dataset as logical facts for symbolic reasoning. Specifically, neural networks are employed to process simple concepts or unstructured data, generating inputs for symbolic reasoning in DeepProbLog. Symbolic reasoning in DeepProbLog utilizes SDD (Sentential Decision Diagrams) \cite{darwiche2011sdd} to construct a directed graph, which is then transformed into an arithmetic circuit for inference and answering queries. To enable end-to-end training that bridges continuous embedding and discrete symbols, DeepProbLog leverages the gradient semiring \cite{eisner2002parameter} as an optimization tool. The framework of DeepProbLog is visually depicted in Figure \ref{fig:deepprolog}.


In a different approach to DeepProbLog, Zhou et al.  \cite{zhou2019abductive} proposed abductive learning (\textbf{ABL}) as a framework that combines abductive reasoning \cite{2006Abductive} with induction. Abductive reasoning, which is a form of logical reasoning, involves inferring the best explanation for given observations or evidence. ABL leverages both induction, which is a key component of modern machine learning, and abduction, which is the process of generating hypotheses or explanations, in a mutually beneficial way. ABL provides a unified framework that bridges machine learning and logical reasoning, allowing for the integration of both approaches to improve the overall learning and reasoning process. This framework offers a new perspective and methodology for effectively combining machine learning techniques with logical reasoning techniques.

In more detail, given raw data (This raw data only includes the data features and a label of true or false, with no label of the class.), an initialized classifier, and a knowledge base (KB), the raw data is fed into the initialized classifier to obtain pseudo-labels in machine learning. These pseudo-labels (pseudo-grounding) are then transformed into symbolic representations that can be accepted by logical reasoning. Next, ABL uses ProLog as the KB and adopts abductive reasoning technology to abduct the pseudo-labels and rules. That is to say, logical reasoning minimizes the inconsistency between the symbolic representation and the KB to revise pseudo-labels, then outputs the deductive labels. Finally, a new classifier is trained by the deductive labels and the raw data, which replaces the original classifier. The above is an iterative process that continues until the classifier is no longer changed or the pseudo-labels are consistent with the KB. ABL is a special kind of weakly supervised learning, in which the supervision information comes not only from the ground-truth labels but also from knowledge abduction. 

Based on the ABL framework, Tian et al. \cite{tian2022weakly} proposed a weakly supervised neural symbolic learning model (\textbf{WS\mbox{-}NeSyL}) for cognitive tasks with logical reasoning.  The difference between {WS\mbox{-}NeSyL and ABL is that ABL uses a metric of minimal inconsistency in logical reasoning, while WS\mbox{-}NeSyL adopts sampling technology. In WS\mbox{-}NeSyL, to provide supervised information for the reasoning process in complex reasoning tasks, the neural network is designed as an encoder-decoder framework that includes an encoder and two decoders (perceptive decoder and cognitive decoder). The encoder can encode input information as a vector, while the perceptive decoder decodes the vector to predict labels (pseudo-labels). According to these pseudo\mbox{-}labels and the sampled logic rules from the knowledge base, the cognitive decoder to reason results. To supervise the reasoning of the cognitive decoder, WS\mbox{-}NeSyL provides a back search algorithm to sample logic rules from the knowledge base to act as labels that are used to revise the predicted labels. To solve the sampling problem, WS\mbox{-}NeSyL introduces a regular term of logic rules. The whole model is trained iteratively until convergence.


The knowledge base is an important factor in logical reasoning, and different knowledge bases are used by different reasoning technologies. The above approaches use probabilistic logic programming language (ProbLog) as their knowledge base; notably, they only consider that neural networks can provide facts for the knowledge base, and do not quantify how many logic rules should be triggered by the neural networks. To resolve this issue, Yu et al. \cite{yu2022probabilistic} proposed a bi-level probabilistic graphical reasoning framework, called \textbf{BPGR}. To quantify the amount of symbolic knowledge that is triggered, BPGR uses MLN to model all logic rules. For instance, MLN can express the time at which a logic rule is true in the form of a potential function.

BPGR includes two parts: the {\em visual reasoning} module ({\em VRM}) and the {\em symbolic reasoning} module({\em SRM}). {\em VRM} extracts the features of objects in images and the inferred labels of objects and relationships. {\em SRM} uses symbolic knowledge to guide the reasoning of {\em VRM} in a good direction, which acts as an error correction. In terms of the model framework, more concretely, {\em SRM} is a double-layer probabilistic graph that contains two types of nodes: one is the reasoning results of the VRD model in the high-level structure, and the other is the ground atoms of logic rules in the low-level structure. When the probabilistic graphical model is constructed, BPGR can be efficiently trained in an end-to-end manner by the variational EM algorithm. An overall framework of BPGR is provided in Figure \ref{fig:BPGR}. 
\begin{figure}[htbp]
\centering
\includegraphics[width=0.8\linewidth]{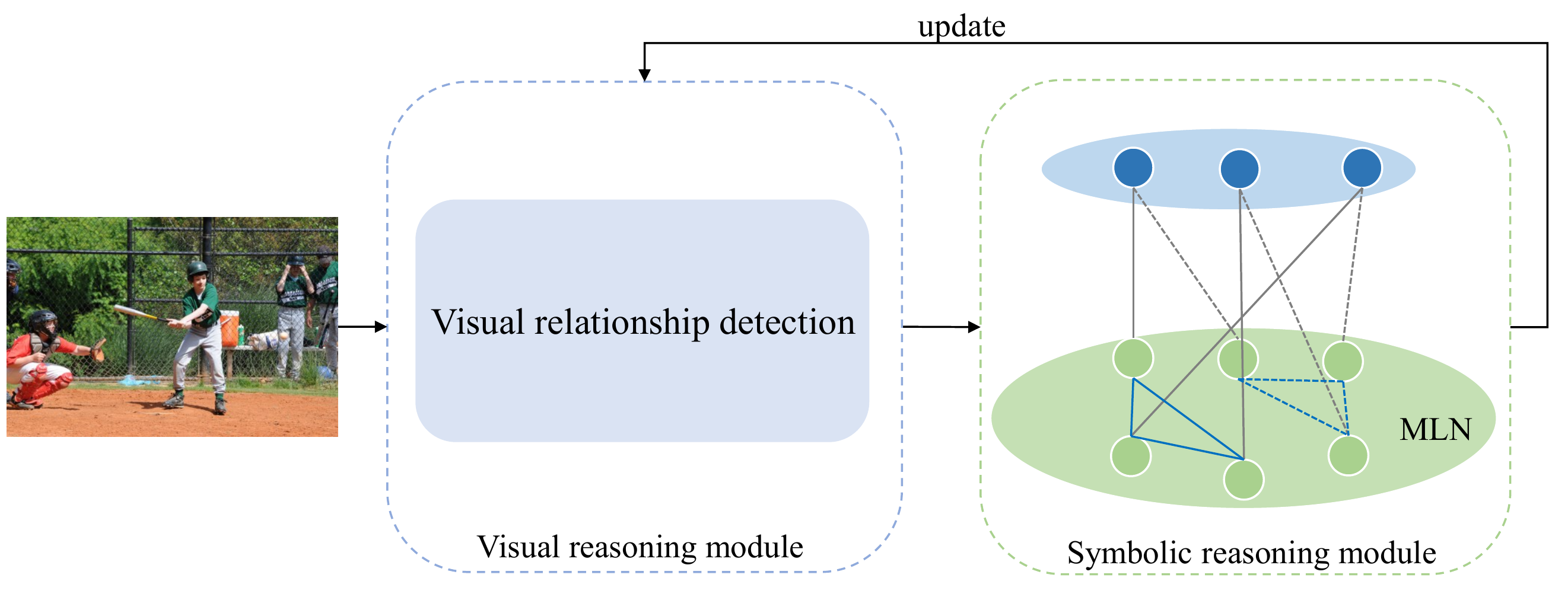}
\caption{The framework of the BPGR. This model is a two-layers probabilistic graphical model which consists of a visual reasoning model and a symbolic reasoning module. Here, The high-level structure is the result of the visual reasoning module, while the low-level structure is the ground atom of logic rules. The model is trained to output reasoning results of the visual reasoning module based on symbolic knowledge. Note that G represents the grounding operator, the solid line represents ground truth edges, and the dotted line represents pseudo-edges.}
\label{fig:BPGR}
\end{figure}

\textbf{Conclusion:} The field of {\em learning\mbox{-}reasoning} approaches in AI research has gained significant attention due to the advantages it offers by combining neural networks and symbolic reasoning. The integration of neural networks and symbolic reasoning allows for the utilization of the strengths of both approaches. Neural networks provide the ability to process complex data and generate predictions, while symbolic reasoning provides a structured and interpretable framework for representing and reasoning about knowledge. For instance, DeepProbLog and ABL have similar model principles: the modeling of complex problems is defined in a logic programming language, and the neural network is used to define simple concepts in a logic programming language. BPGR uses neural networks to accelerate the search process of symbolic reasoning, along with symbolic knowledge to constrain neural network learning. This model not only characterizes the matching degree between prediction results and symbolic knowledge but also clearly states which symbolic knowledge is being fitted, along with the probability of this symbolic knowledge being fitted as an explanation for the model prediction. However, one limitation of these approaches is their reliance on predefined logic programming or logic rules, which restricts their generalizability to other tasks. Future research in this field should explore higher-level interactions between neural networks and symbolic reasoning, such as learning symbolic knowledge during training. By enhancing the ability of models to acquire and reason with symbolic knowledge in a data-driven manner, the field can further advance the integration of learning and reasoning in AI systems.

Based on representative works, we summarize a general design idea for neural-symbolic approaches. The interaction between neural networks and symbolic systems allows encoding the embedding of symbolic knowledge into neural network models and feeds abstracted symbols of the neural networks into symbolic systems. Further, we provide some characteristics that should be considered in designing neural-symbolic approaches, as follows: (1) \emph{Uncertainty}. The output of the neural network is a distribution, not "True" or "False". Therefore, we need to consider the uncertainty of the triggered symbolic knowledge. (2) \emph{Globalization}. It is necessary to consider the fit of all symbolic knowledge in the knowledge base, not just the local knowledge. (3) \emph{Importance}. Different knowledge may have different weights, and the degree of fitting knowledge with different weights should be considered. (4) \emph{Interpretability}. Interpretability should be explicitly considered in learning (e.g. the immediate process of the result of learning).
\section{Applications}
\label{sec5}
\subsection{Object/visual-relationship detection}
 The goal of object/visual-relationship detection is to recognize objects or the relationships between objects in images. However, relying solely on visual features to train a model often leads to relatively weak performance. In recent years, the emergence of neural-symbolic learning systems has paved the way for incorporating external knowledge to enhance the detection performance of these models. This integration of external knowledge into the learning process has shown promising results and has become an active area of research in the field.

Donadello et al. \cite{donadello2017logic} propose a novel approach that combines neural networks with first-order logic, known as Logic Tensor Networks (LTN). By incorporating logical constraints, LTNs enable effective reasoning from noisy images while also providing a means to describe data characteristics through logic rules. This integration of logic into the neural network framework enhances interpretability in image recognition tasks. In the context of remote sensing, Marszalek and Forestier et al. \cite{marszalek2007semantic,forestier2013coastal} emphasize the utilization of symbolic knowledge from domain experts to improve the detection capabilities. By leveraging expert knowledge, the remote sensing systems can gain a deeper understanding of the data and achieve better performance in detecting specific features or patterns. Zhu and Nyga et al. \cite{zhu2014reasoning,nyga2014pr2} adopt a different approach by using Markov Logic Networks (MLN) to model symbolic knowledge for integration into deep learning models. MLNs allow for learning a scoring function and predicting relations between input images and specific objects or concepts. For example, given an input image of a horse, the model can predict the relation "ridable" between the horse and people. This approach combines the strengths of deep learning and symbolic reasoning, enabling more comprehensive and nuanced analysis of visual data. 

\subsection{Knowledge graph reasoning}
Knowledge graphs often suffer from incompleteness, requiring completion or link prediction techniques to enhance their quality. Zhang et al. \cite{zhang2021neural} provide a comprehensive survey on the benefits of knowledge graph reasoning within neural-symbolic learning systems. Wang et al. \cite{wang2019logic} propose a method where triplets or ground rules are transformed into First-Order Logic (FOL) statements. These FOL statements are then scored using vector/matrix operations based on the embeddings of the entities and relationships involved. This approach combines logic reasoning with neural networks to perform link prediction in knowledge graphs. Path-based reasoning approaches \cite{neelakantan2015compositional, das2016chains, xiong2017deeppath, meilicke2020reinforced, das2017go} aim to extend reasoning by exploring multi-hop neighbors around a given entity and predicting answers within these neighborhoods using neural networks. For example, DeepPath \cite{xiong2017deeppath} employs reinforcement learning to evaluate sampled paths, reducing the search space and improving efficiency. Building upon path-based reasoning, Teru et al. \cite{teru2020inductive} propose GraIL, a graph-based reasoning framework. GraIL extracts a subgraph consisting of the k-hop neighbors of the head and tail entities. Subsequently, a Graph Neural Network (GNN) is employed to reason about the relationship between the two entities using the extracted subgraph. These approaches demonstrate the integration of neural networks and graph-based reasoning to perform link prediction and enhance knowledge graph reasoning.

\subsection{Classification/ few-shot classification} 
Marra et al. \cite{marra2020relational} introduce Relational Neural Machines (RNM), a framework that allows for joint training of learners and reasoners. This approach enables the integration of both learning and reasoning processes, leading to improved performance. To address the few-shot learning problem, Sikka et al. \cite{sikka2020zero} incorporate common sense knowledge into deep neural networks. They also utilize logical knowledge as a neural-symbolic loss function to regularize visual semantic features. This approach leverages information from unseen classes during model learning, enhancing zero-shot learning capabilities. Altszyler et al. \cite{altszyler2020zero} integrate logic rules into neural network architectures for multi-domain dialogue recognition tasks. By incorporating logical knowledge, the model can recognize labels of unseen classes without requiring additional training data. This approach expands the model's ability to handle previously unseen classes in the dialogue recognition domain. These methods demonstrate the integration of logical knowledge, common sense knowledge, and neural networks to improve the performance of various tasks, such as joint training, few-shot learning, and multi-domain dialogue recognition.

\subsection{Intelligent question answering} 
Indeed, intelligent question answering is a prominent application of neural-symbolic reasoning in natural language processing and visual reasoning tasks. In this context, the goal is to develop models that can accurately infer answers to questions by leveraging contextual information from text and images.

Andreas et al. \cite{andreas2016neural} proposed the neural module network (NMN) framework, which uses deep neural network generating symbolic structures to solve subsequent reasoning problems. Gupta et al. \cite{gupta2019neural} extend NMN and propose an unsupervised auxiliary loss to help extract arguments associated with the events in the text. Specifically, this method introduces a reasoning module for the text that enables symbolic reasoning (such as arithmetic, sorting, and counting) on numbers and dates in a probabilistic or differentiable way, allowing the model to output logical parsing of questions and intermediate decisions.

Hudson et al. \cite{hudson2018compositional} propose a fully differentiable network model (MAC) with cyclic memory, attention, and composition functions. The MAC provides strong prior conditions for iterative reasoning, transforms the black-box architecture into something more transparent, and supports interpretability and structured learning. The core concept is to decompose the image and the question into sequential units, input the recurrent network for sequential reasoning, and then store the result in the memory unit to calculate the final answer together with the question. Tran and Poon et al. \cite{tran2008event,poon2009unsupervised} model domain common sense with MLN and use probabilistic inference methods for the query. Sun et al. \cite{sun2020neural} learned a neural semantic parser and trained a model-agnostic model based on meta-learning to improve the predictive ability of language question-answering tasks cases involving limited simple rules.

Oltramariet et al. \cite{oltramari2021generalizable} proposed integrating neural language models and knowledge graphs in common-sense question answering. Based on the architecture of the language model, this work proposes an attention-based knowledge injection method. For visual question-answering tasks, Hudson et al. \cite{hudson2019learning} propose the neural state machine (NSM). NSM uses a supervised training method to construct a probabilistic scene graph based on the concepts in an image, then performs sequential reasoning on the probabilistic scene graph, answering questions or discovering new conclusions.

\subsection{Reinforcement learning} 
Deep reinforcement learning is a trending topic in the field of artificial intelligence, and it has been successfully applied in many contexts. However, current deep reinforcement learning methods have limitations in reasoning ability. To address this challenge, researchers have started integrating symbolic knowledge into reinforcement learning. In this paper, we explore two approaches: combining symbolic knowledge with deep reinforcement learning and combining symbolic knowledge with hierarchical reinforcement learning. 

Garnelo et al. \cite{garnelo2016towards} proposed a deep symbolic reinforcement learning method (DSRL) that integrates a symbolic prior into the agent's learning process to enhance the model's generalization. The DSRL agent consists of a neural back end and a symbolic front end. The neural back end learns to map raw sensor data into a symbolic representation, which is then utilized by the symbolic front end to learn an effective policy. Building upon DSRL, Garcez et al. \cite{garcez2018towards} extended the approach and introduced a symbolic reinforcement learning method with common sense (SRL+CS). This method improves both the learning phase and the decision-making phase based on DSRL. In the learning phase, the reward distribution is no longer based on a fixed calculation formula for updating the Q-values, but instead considers the interaction between the agent and the object. In the decision-making stage, the model assigns an importance weight to each Q-function based on the distance between the object and the agent, allowing for a comprehensive aggregation of the Q-values.

Yang et al. \cite{yang2018peorl} proposed the integration of symbolic planning and hierarchical reinforcement learning (HRL) \cite{barto2003recent} to address decision-making in dynamic environments with uncertainties. They introduced a framework called PEORL (Planning-Execution-Observation-Reinforcement-Learning) that combines these two approaches. Symbolic planning is employed to guide the agent's task execution and learning process, while the learned experiences are fed back to the symbolic knowledge to enhance the planning phase. Specifically, commonsense knowledge of actions constrains the answer set solver to generate a symbolic plan. The symbolic plan is subsequently mapped to a deterministic sequence of stochastic options, which guides the hierarchical reinforcement learning (HRL) process. This approach represents the first utilization of symbolic planning for option discovery within the HRL framework. To achieve task-level interpretability, Lyu et al. \cite{lyu2019sdrl} proposed the Symbolic Deep Reinforcement Learning (SDRL) framework, which shares similarities with REORL and consists of a planner, controller, and meta-controller, along with symbolic knowledge. The planner employs prior symbolic knowledge to perform long-term planning through a sequence of symbolic actions (subtasks) that aim to achieve its intrinsic goal. The controller utilizes deep reinforcement learning (DRL) algorithms to learn sub-policies for each subtask based on intrinsic rewards. The meta-controller learns extrinsic rewards by evaluating the training performance of the controllers and suggesting new intrinsic goals to the planner. In essence, both PEORL and SDRL leverage symbolic knowledge to guide the reinforcement learning process and facilitate decision-making.

\section{Future directions}
\label{sec6}
The above paper introduces the current research status and research methods of neural-symbolic learning systems in detail. On this basis, we discuss some potential future research directions.

\subsection{Efficient methods}
In neural-symbolic learning systems, symbolic reasoning technologies often encounter challenges related to intractable precise inference. For instance, in probability inference using Markov Logic Networks (MLNs), the number of groundings can increase exponentially when dealing with a large number of logic rules and constants. This exponential growth in grounding can significantly decrease the speed of model inference. While several methods have been proposed to mitigate this problem, such as learning-based approaches \cite{khot2011learning, bach2015hinge}, they still have certain limitations. Approximate inference techniques are commonly employed to improve inference speed but often come at the cost of reduced accuracy. Consequently, it becomes crucial for researchers to explore the potential of neural networks to tackle the computational challenges faced by symbolic systems. Designing approaches that leverage the computational strengths of neural networks to handle tasks that are computationally difficult in traditional symbolic systems represents a crucial research direction for advancing inference methods.

\subsection{Automatic construction of symbolic knowledge}
The symbolic knowledge discussed in this paper encompasses logic knowledge and knowledge graphs. Automatic construction of symbolic knowledge has achieved relative maturity, as evidenced by studies on automatic knowledge construction \cite{liuqiao2016knowledge, luan2018multi, martinez2018openie}. However, the construction of logic rules in neural-symbolic approaches typically relies on manual efforts from domain experts. This process is time-consuming, laborious, and not easily scalable. A significant challenge for neural-symbolic learning systems is enabling end-to-end learning for rules that describe prior knowledge derived from data. While technologies like Inductive Logic Programming (ILP)-based methods exist for knowledge extraction, the automatic learning of logic rules from data remains largely unexplored. Hence, the automatic construction of rules represents an important future research direction in the field of neural-symbolic learning systems.

\subsection{Symbolic representation learning}
A well-designed symbolic representation plays a crucial role in simplifying and improving the efficiency of complex learning tasks. In the context of zero-shot image classification, for instance, a limited semantic information contained in a learned symbolic representation can hinder the model's ability to handle complex classification tasks effectively. Therefore, precise semantic information encoded in symbolic knowledge is essential for enhancing the performance of these models. However, most existing symbolic representation learning methods struggle with predicates that exhibit strong similarity. These predicates may have similar semantics but different logical formulas, such as "next to" and "near." Current symbolic representation learning methods fail to capture such semantic similarities, which hampers the reasoning capability of these models. Thus, a significant challenge in the field of neural-symbolic learning systems is the design of more robust and efficient symbolic representation learning methods. The advancement of graph representation learning offers a promising avenue for addressing this challenge. By mapping nodes to low-dimensional, dense, and continuous vectors, graph representation learning can flexibly support various learning and reasoning tasks. Given that symbolic knowledge often exhibits heterogeneity, multiple relations, and even multimodality, exploring the development and utilization of heterogeneous graph representation learning methods becomes another important direction to overcome the challenges faced by neural-symbolic learning systems.

\subsection{Application field expansion}
Neural-symbolic learning systems have found applications in various domains, including computer vision, natural language processing, and recommendation systems \cite{zhu2021faithfully,raizada2022survey}. Recently, researchers have also started exploring the potential of applying neural-symbolic learning systems in other areas such as the COVID-19 pandemic \cite{ngan2022extracting} and advanced robotics \cite{9216410}. For example, in the context of the COVID-19 pandemic, neural-symbolic learning systems have been utilized for tasks like extracting relevant information from medical literature. Similarly, in the field of advanced robotics, neural-symbolic learning systems can be employed to enhance robot intelligence and decision-making capabilities. Given the success of neural-symbolic learning systems in existing domains, it is a natural and promising direction to extend their application to new domains and develop tailored methods to address specific challenges in those areas.

\section{Conclusion}
\label{sec7}
In this paper, we have presented an overall framework for neural-symbolic learning systems. Our main contribution is the proposal of a novel taxonomy for neural-symbolic learning systems and the outline of three structured categorizations. Additionally, we describe the techniques used in each structured categorization, explore a wide range of applications, and discuss future directions for neural-symbolic learning systems. We firmly believe that a systematic and comprehensive research survey in this field holds significant value in terms of both theory and application. It deserves further in-depth research and discussion.

\section*{Acknowledgments}
This research was partly supported by Foundation item: National Natural Science Foundation of China (61876069); Jilin Province Key Scientific and Technological Research and Development Project under Grant Nos. 20180201067GX and 20180201044GX; and Jilin Province Natural Science Foundation (20200201036JC).



\printcredits

\bibliographystyle{cas-model2-names}

\bibliography{cas-refs}

\appendix
\section{Preliminaries}
\label{sec2}
In this section, we introduce background information related to symbolic knowledge and neural networks. For symbolic knowledge, we focus on two categories: logic knowledge and knowledge graphs. Logic knowledge can be further subdivided into propositional logic and first\mbox{-}order logic.
\subsection{Symbols}
\subsubsection{Propositional logic}
Propositional logic statements are declarative sentences that are either True or False. A declarative sentence is a True sentence if it is consistent with the fact; otherwise, it is a False sentence. The connectors between propositions are  "$\wedge$", "$\vee$", "$\lnot$" and "$\Rightarrow$". Propositional logic can be expressed in the form of the following formula:
\begin{equation}\label{eq:2}
    P \Rightarrow Q,
\end{equation}
where $P$ represents the antecedent (condition), while $Q$ represents the consequent (conclusion).

Propositional logic is usually compiled in the form of directed acyclic graphs. Conjunctive Normal Forms (CNFs), deterministic-Decomposable Negation Normal Forms (d-DNNFs) \cite{darwiche2001tractable,darwiche2002knowledge}, and Sentential Decision Diagrams (SDDs) \cite{darwiche2011sdd} are representative examples of
knowledge representation, where SDD is a subset of d-DNNFs. For example, given a propositional logic $Smokes \Rightarrow Cough$, its CNF and d-DNNF graph are presented in Figure \ref{fig:CNF} (a) and Figure \ref{fig:CNF} (b), respectively.

\begin{figure}[h]
\centering
\includegraphics[width=0.8\linewidth]{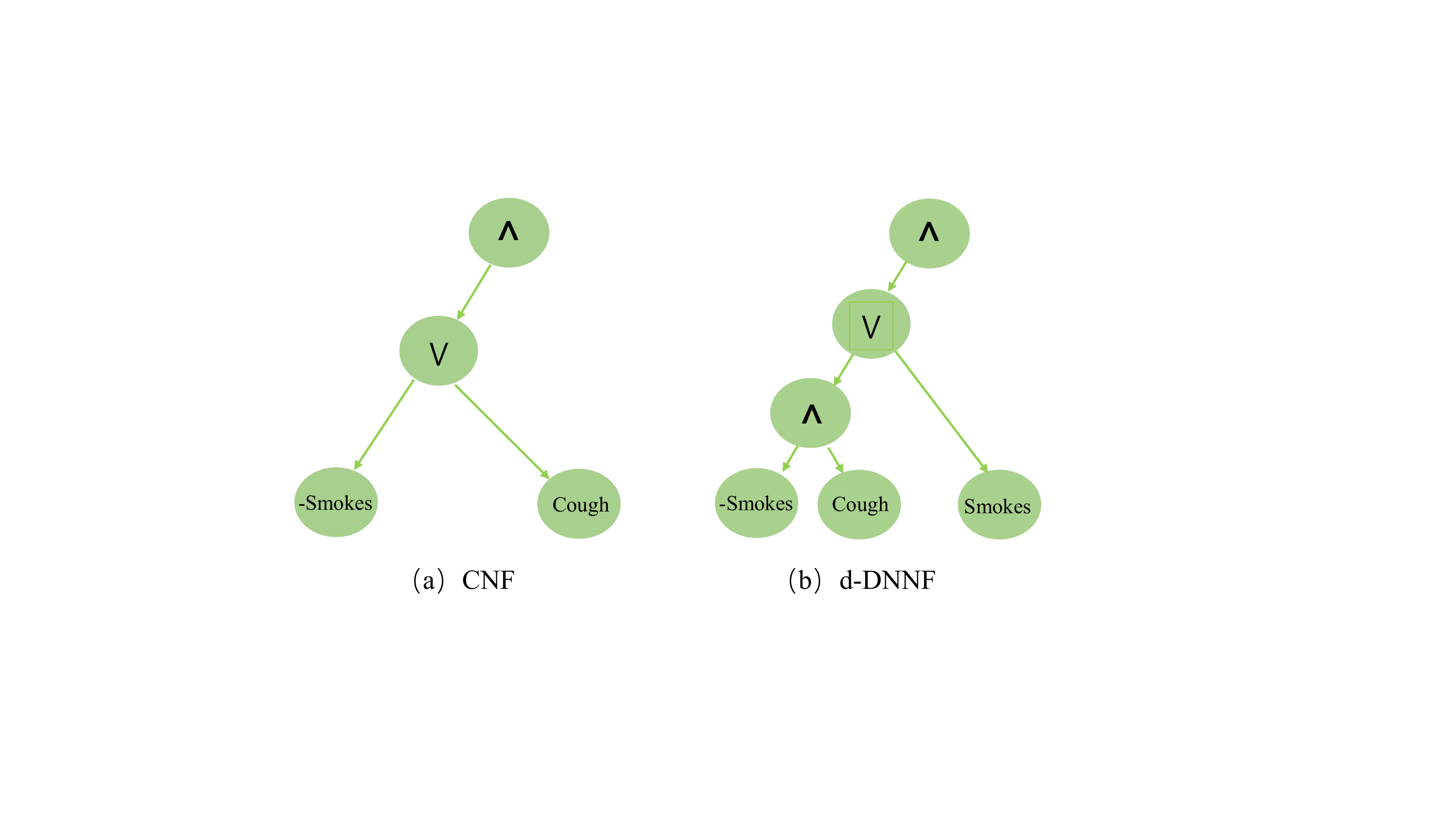}
\caption{Directed acyclic graph of CNF and d-DNNF. In the graph, leaf nodes represent atoms of propositional logic, and other nodes represent connectors. Directed edges are the relationship between nodes.}
\label{fig:CNF}
\end{figure}

\begin{table*}[pos=!h]\footnotesize
\renewcommand{\arraystretch}{1.3}
\centering
\caption{An instance of a Markov logical network.}
\label{tab:4}
\begin{tabular}{ccc}
\hline
\textbf{Proposition}&\textbf{First\mbox{-}order logic}&\textbf{Weight}\\ 
\hline
Smoking causes cough. & $F1: {\forall}x, Smokes(x) \Rightarrow Cough(x)$ & 1.5\\

If two people are friends, either both smoke or neither does. & $F2: {\forall} x {\forall} y, Friends(x,y) \Rightarrow (Smokes(x) \Leftrightarrow Smokes(y)) $& 1.1 \\
\hline
\end{tabular}
\end{table*}

\subsubsection{First\mbox{-}order logic}
Propositional logic is hard to describe complex problems; for these, predicate logic is required. In this paper, we introduce one of the predicate logic --first\mbox{-}order logic (FOL) \cite{enderton2001mathematical}. FOL consists of four types of elements, connectors, and quantifiers. The four types include constants, variables, functions, and predicates. Constants represent objects in the domain of interest (for example, in the predicate father(a,b), a=Bob, b=Mara, a and b are constant). Variables range over the objects in the domain (for example, in the predicate father($x$,$y$), where $x$ is the father of $y$, and the variable $x$ is limited to the scope of the father class). Functions represent mappings from tuples of objects to objects. Predicates represent relations among objects in a given domain or attributes of these objects. The connector is the same as in propositional logic. FOL involves the combination of atoms through connectors, such that an expression can be written in the following form:
\begin{equation}\label{eq:1}
    B_1(x)\wedge B_2(x)\wedge \cdots \wedge  B_n(x)\Rightarrow H(x),
\end{equation}
where $B_1(x)$, $B_2(x)$, $\cdots$, $B_n(x)$ represents the rule body, which is composed of multiple atoms. $H(x)$ represents the rule head and is the result derived from the rule body.

Knowledge representation of FOL can be achieved by a Markov logic network (MLN) \cite{richardson2006markov}. MLN is an undirected graph in which each node represents a variable, and the joint distribution is represented as follows: 
\begin{equation}\label{eq:10}
    P(X=x)= 1/Z exp\{\sum_i w_i n_i(x)\},
\end{equation}
where $Z$ represents the partition function, $w_i$ represents the weight of the rule, $n_i(x)$ represents the number of times that the value of the rule is true, and we use t-norm fuzzy logic \cite{novak2012mathematical} to calculate logical connectives. 

The following introduces a simple example of an MLN. Table \ref{tab:4} shows the two rules ($F_1$,1.5), ($F_2$,1.1) of this example \cite{domingos2019unifying}. Given a constant set $C = \{A, B\}$, the generated ground Markov logic network is shown in Figure \ref{fig:9}.

\begin{figure}[htbp]
\centering
\includegraphics[width=0.8\linewidth]{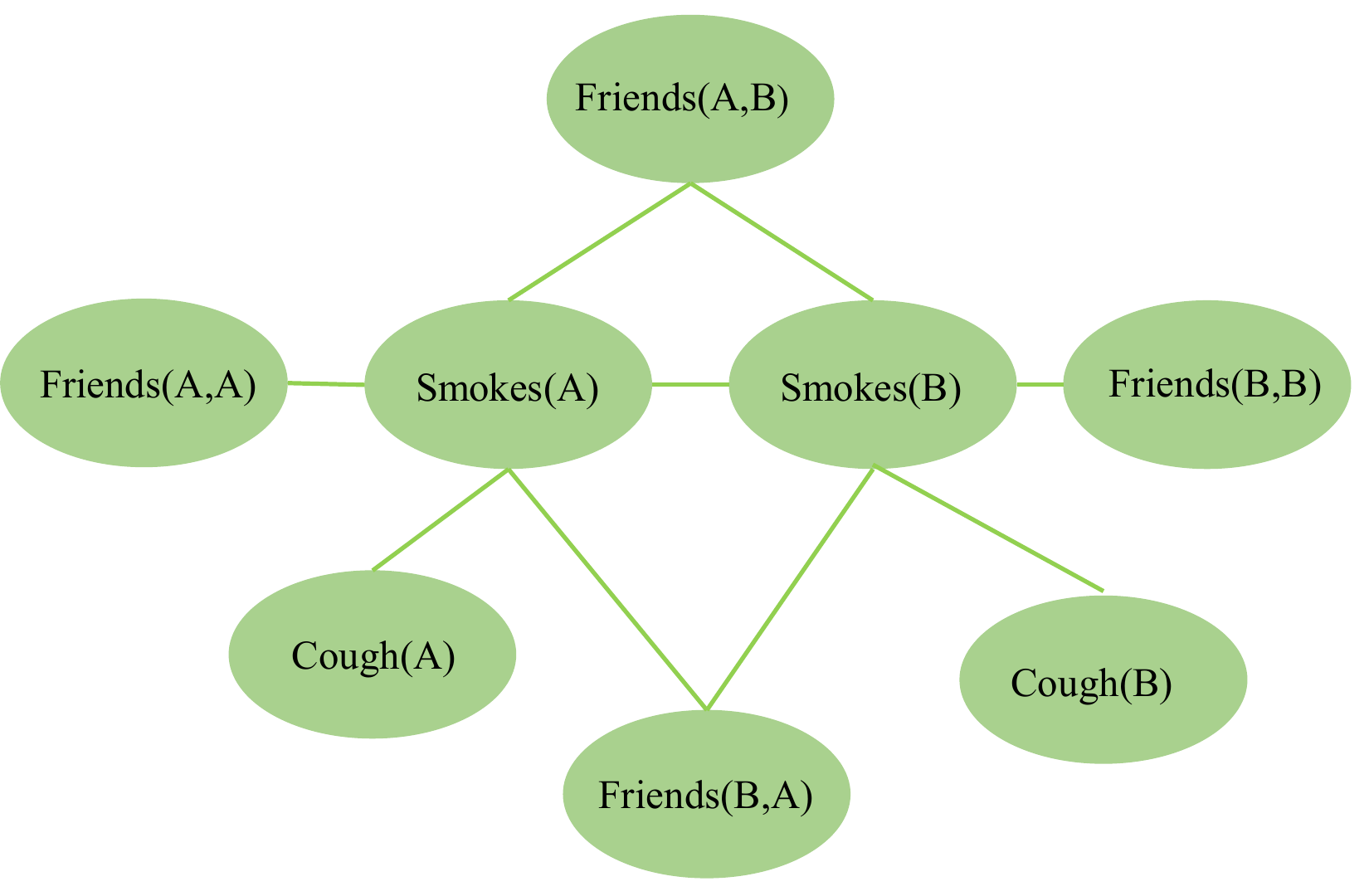}
\caption{Ground Markov logic network. Nodes are variables and edges are the relationship between variables.}
\label{fig:9}
\end{figure}

\subsubsection{ Knowledge graph}
A knowledge graph is a directed and labeled graph. Nodes in this graph represent semantic symbols (concepts), such as animals, computers, people, etc; for their part, edges connect node pairs and express the semantic relationships between them, such as the food chain relationship between animals, friend relationship between people, etc. Knowledge graphs can be formally expressed in the form: $G = (H, R, T)$, here, $H = \{ h_1, h_2,..., h_n\}$ represents the set of head entities, and $n$ represents the number of head entities, $T = \{t_1, t_2,..., t_m\}$ represents the set of tail entities, and $m$ represents the number of tail entities; moreover, $R = \{r_1, r_2,..., r_k\}$ represents the set of relationships, and $k$ represents the number of relationships. Figure \ref{fig:7} presents an example. For a given triplet (cat, attribute, paw), the nodes (head entity and tail entity) are cat and paw, and the relationship is an attribute.

\begin{figure}[htbp]
\centering
\includegraphics[width=0.8\linewidth]{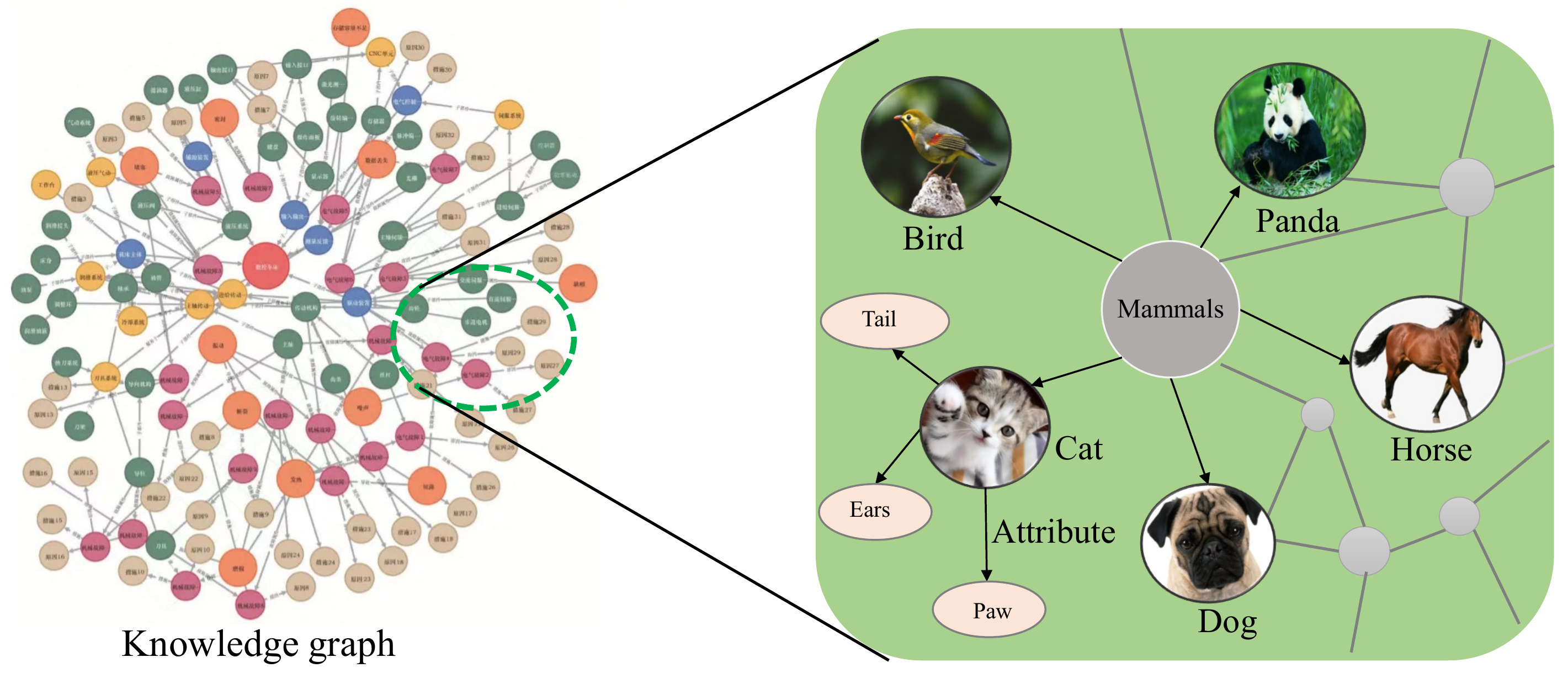}
\caption{An example of a knowledge graph.}
\label{fig:7}
\end{figure}

For their part, a knowledge graph representation is used to encode discrete symbols (entities, attributes, relationships, etc.) into a low-dimensional vector space to obtain a distributed representation. Typical methods include R-GCN \cite{schlichtkrull2018modeling}, M-GNN \cite{wang2019robust}, CompGCN \cite{vashishth2019composition}, TransE \cite{bordes2013translating},TransR \cite{lin2015learning}, TransH \cite{wang2014knowledge}, RotatE \cite{sun2019rotate}, DisMult \cite{yang2014embedding}, ComplEX \cite{trouillon2016complex}, ConvE \cite{dettmers2018convolutional}, ConvR \cite{jiang2019adaptive}, GGNN \cite{li2015gated}, and GCN \cite{kipf2016semi}, among others.

\subsection{Neural networks}
Although neural networks exist in many forms, we only
introduce those relevant to works discussed in Section \ref{sec4}.
\begin{figure}[h]
\centering
\includegraphics[width=0.7\linewidth]{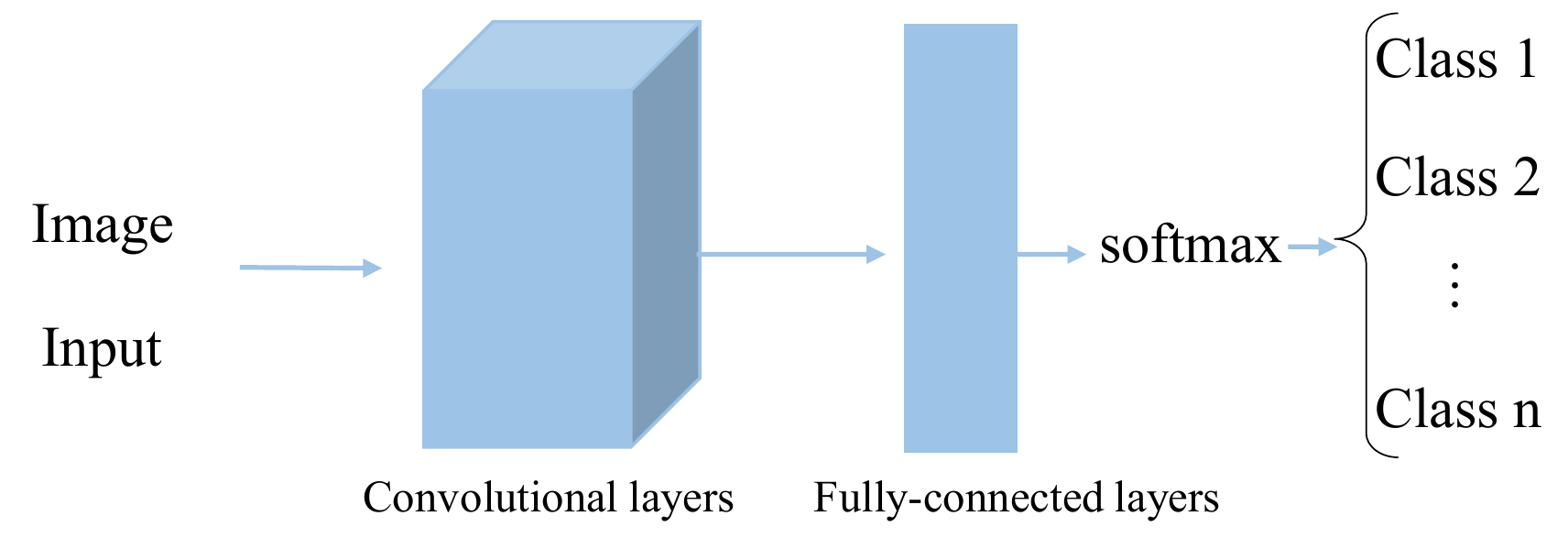}
\caption{Diagram of a CNN.}
\label{CNN}
\end{figure}

CNNs (Convolution Neural Networks, Figure \ref{CNN}) \cite{lecun1998gradient} have a set of convolutional layers that extracts features from the input (The input usually is the image). Next, one or more fully-connected layers are applied to extract the classification or regression information with the help of logistic or linear regression. Usually, the final layer of a CNN is a "softmax" layer, where the sum of all activations is one.
\begin{figure}[h]
\centering
\includegraphics[width=0.8\linewidth]{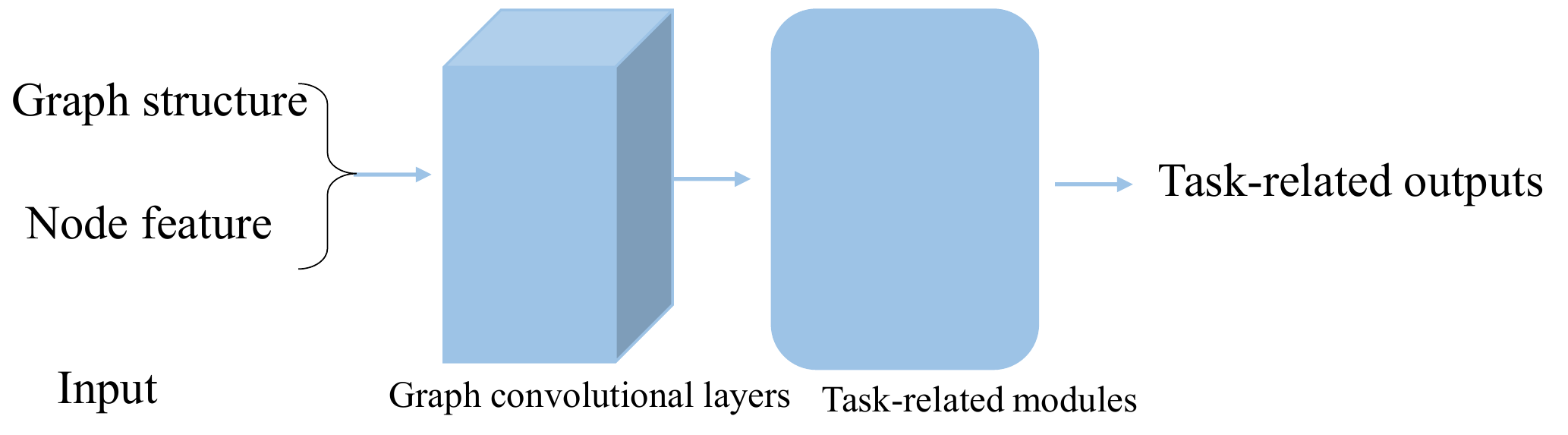}
\caption{GNN, which feeds graph structure and node feature into graph convolutional layers.}
\label{GNN}
\end{figure}

GNNs (Graph Neural Networks, Figure \ref{GNN}) \cite{wu2020comprehensive} with multiple graph convolutional layers that can attain each node’s hidden representation by aggregating feature information from its neighbors. The graph structure and node feature information as inputs and the outputs of GNNs can focus on different graph analysis tasks, such as node classification and graph classification. Therefore, GNNs have different variants, such as GCN, GAE, etc.

\begin{figure}[h]
\centering
\includegraphics[width=0.5\linewidth]{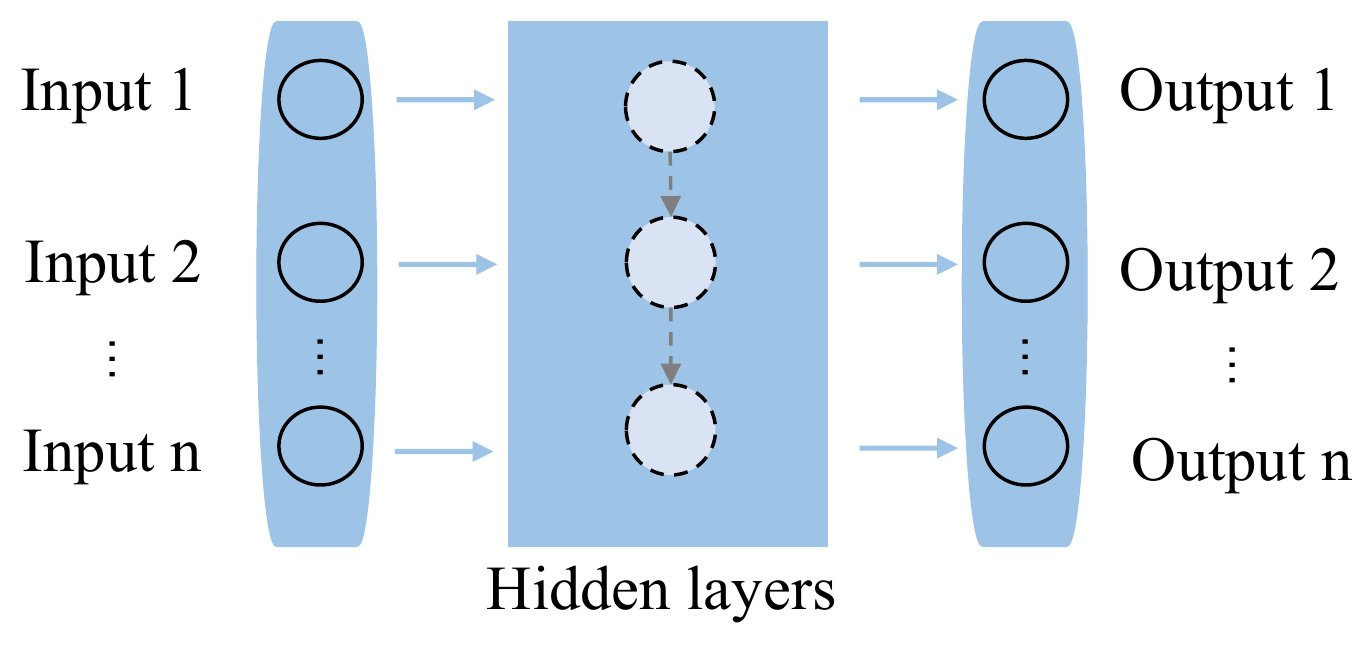}
\caption{RNN, which feeds its output back into its input. Here, dashed lines represent recurrent connections.}
\label{RNN}
\end{figure}

RNNs (Recursive Neural Networks, Figure \ref{RNN}) \cite{hoffmann2017survey} are neural networks in which the input and output layers are identical, and the latter propagates activation back to the former through recurrent connections. Over time, with each time step consisting of a full input-to-output forward pass, the hidden layer transitions through a series of states. Usually, the input is sequence data such as sentences in natural language processing. LSTM (Long Short-Term Memory Networks) is a variable of RNN.
\bio{}
\endbio

\bio{}
\endbio

\end{document}